\documentclass[journal]{IEEEtran}
\IEEEoverridecommandlockouts

\usepackage[english]{babel}
\usepackage{microtype}
\usepackage{amsmath,amsfonts}
\usepackage{algorithmic}
\usepackage{algorithm}
\usepackage{array}
\usepackage[caption=false,font=normalsize,labelfont=sf,textfont=sf]{subfig}
\usepackage{textcomp}
\usepackage{stfloats}
\usepackage{url}
\usepackage{verbatim}
\usepackage{graphicx}
\usepackage{cite}
\usepackage{amsmath}
\usepackage{amssymb}
\usepackage{booktabs}
\usepackage{multirow}
\usepackage{amsfonts}
\usepackage{enumerate}
\usepackage{tabularx}
\usepackage{algorithm,algorithmic}
\usepackage{bm}
\usepackage{enumerate}
\usepackage{adjustbox}
\usepackage{xcolor}
\usepackage{siunitx}
\usepackage{mdframed}
\usepackage{adjustbox}
\usepackage{authblk}
\usepackage{color}
\usepackage{xurl}
\usepackage{subcaption}
\usepackage{cite}
\makeatletter
\let\NAT@parse\undefined
\makeatother
\usepackage{hyperref}
\usepackage{relsize}
\usepackage{float}
\usepackage{pifont}
\usepackage{textcomp}
\usepackage{graphicx}
\usepackage{caption}
\usepackage{subcaption}
\usepackage{pifont}
\usepackage{multirow}
\usepackage{makecell}

\usepackage[most]{tcolorbox}
\definecolor{takeawaygreen}{RGB}{34,139,34}
\definecolor{takeawaybg}{RGB}{240,249,240}

\newtcolorbox{takeawaybox}{
    enhanced,
    colback=takeawaybg,
    colframe=takeawaygreen,
    boxrule=0.7pt,
    arc=1.2mm,
    left=1mm,
    right=1mm,
    top=0.6mm,
    bottom=0.6mm,
    boxsep=0.8mm,
    width=\columnwidth,
    before skip=4pt,
    after skip=6pt
}

\begin{document}

\title{SkyJEPA: Learning Long-Horizon World Models for Zero-Shot Sim-to-Real Control of Quadrotors}

\author{
Pratyaksh Rao$^{1}$,
Wancong Zhang$^{2}$,
Randall Balestriero$^{3}$,
Yann LeCun$^{2}$,
Giuseppe Loianno$^{1}$%
\thanks{$^{1}$The authors are with the University of California Berkeley, Department of Electrical Engineering and Computer Sciences, Berkeley, CA 94720, USA. e-mail: pratyaksh10@berkeley.edu, loiannog@eecs.berkeley.edu}%
\thanks{$^{2}$The authors are with New York University, New York, NY, USA. e-mail: wz1232@nyu.edu, yann@cs.nyu.edu}%
\thanks{$^{3}$The author is with Brown University, Providence, RI, USA. e-mail: randall\_balestriero@brown.edu}%
}


\makeatletter


\makeatother
\maketitle

\begin{abstract}
Accurate dynamics models are critical for informed decision-making in robotic systems, particularly for agile aerial vehicles operating under uncertainty. Neural network dynamics models are attractive for capturing complex nonlinear effects, but existing predictive approaches struggle with long-horizon forecasting because their autoregressive rollout mechanism amplifies errors over time. Joint Embedding Predictive Architectures (JEPAs) offer a compelling alternative by modeling dynamics in latent space, yet prior JEPA-style methods for robot navigation have been studied primarily for kinematic-level planning, with limited investigation in high-frequency control. In this work, we introduce the JEPA-style model for real-time quadrotor control. The proposed approach combines a latent dynamics model with a novel physics-inspired prober that maps frozen latents to interpretable state, enabling physically grounded long-horizon prediction. Additionally, we combine the learned model with a sampling-based optimal control solution to take advantage of its predictive capabilities for real-time control on embedded hardware. Finally, to reduce the dependence on expensive and unsafe real-world data collection, we develop a structured pipeline for automated dataset generation. Extensive open-loop and outdoor closed-loop experiments demonstrate accurate prediction, robust zero-shot sim-to-real transfer, and strong generalization across diverse operating conditions.

\end{abstract}

\begin{IEEEkeywords}
Model Learning for Control; Aerial Systems:
Mechanics and Control; Learning and Adaptive Systems; Opti-
mization and Optimal Control
\end{IEEEkeywords}


\section{Introduction}
\IEEEPARstart{U}nmanned Aerial Vehicles (UAVs) have become increasingly important in applications such as package delivery, infrastructure inspection, search and rescue, and environmental monitoring~\cite{kumar2012opportunities, wang2023wind}. These tasks require aerial robots to operate reliably in complex, uncertain, and often rapidly changing environments while executing agile maneuvers with limited onboard sensing and computation. Achieving this level of autonomy fundamentally depends on the ability to make accurate decisions. For this reason, model-based control has emerged as a particularly attractive paradigm for aerial robotics, as it explicitly reasons about future system evolution and can naturally incorporate task objectives, dynamical constraints, and robustness considerations~\cite{chua2018deep, williams2017information, m2023model}.

\begin{figure}[t]
  \centering
  \includegraphics[width=\linewidth, trim=0 150 350 0, clip]{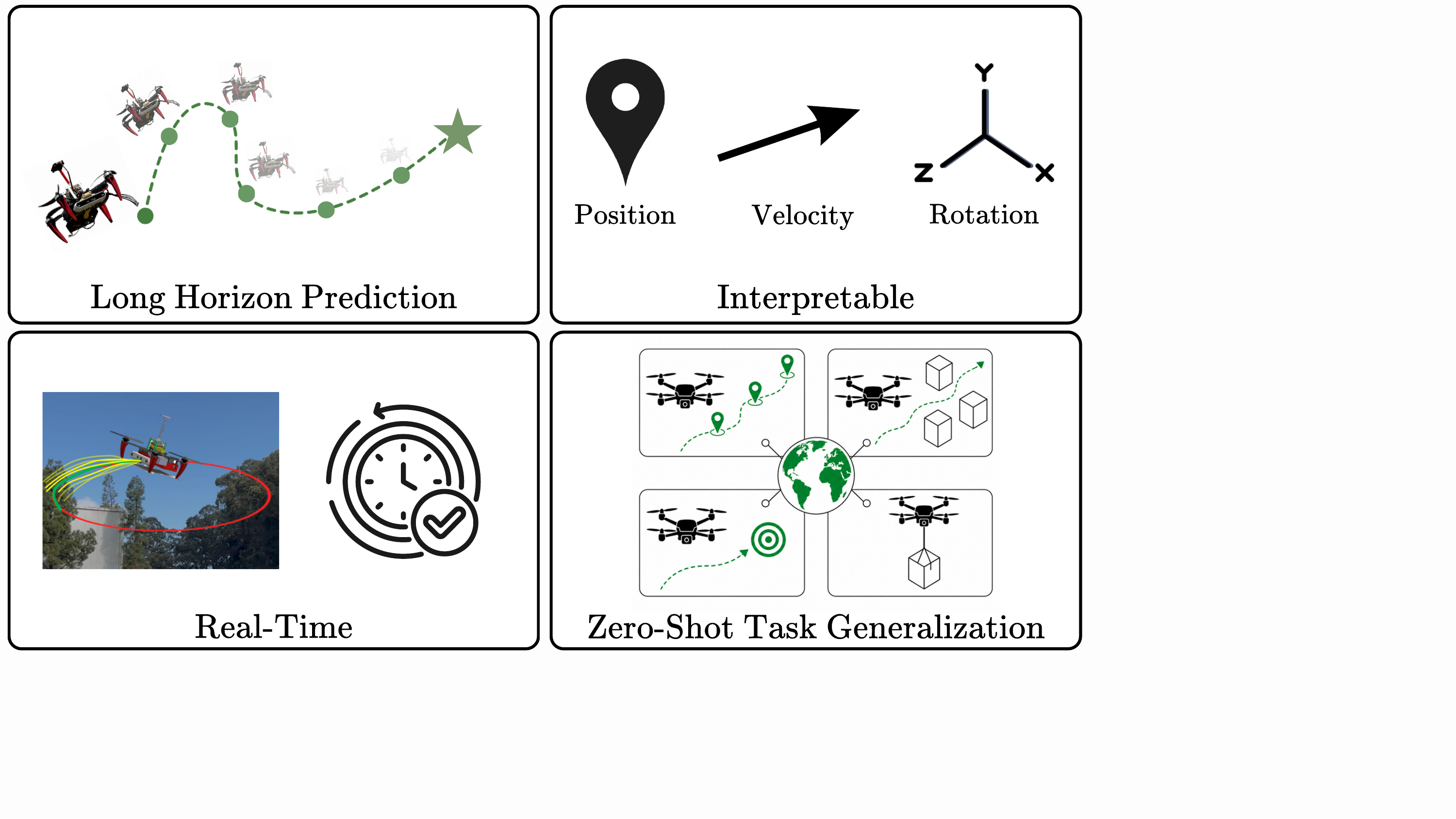}
   \caption{Desirable properties of a quadrotor world model: accurate long-horizon prediction, interpretable, real-time inference for closed-loop control, and zero-shot task generalization.}
   \label{fig:wm_properties}
\end{figure}

A dynamics model, often referred to as predictive or world model, for aerial control should satisfy four key properties (see Figure~\ref{fig:wm_properties}). First, it should provide \emph{accurate long-horizon predictions}. Its rollouts must remain accurate, stable, and physically plausible over the prediction horizons used for control. This requires capturing both the dominant rigid-body dynamics and difficult-to-model effects such as aerodynamic drag, actuator delay, propeller--airframe interactions, wind disturbances, and hardware variations. Second, it should be \emph{interpretable}. The model should be able physically to capture and expose meaningful quantities such as position, velocity, attitude, and angular velocity. This is essential for enforcing constraints, actuator limits, safety bounds, and task costs inside a controller. Third, it should be \emph{real-time}. Real-time aerial control requires repeated high-frequency model evaluations inside resource-constrained onboard optimization loops. Finally, the last property is \emph{zero-shot task generalization}. The same dynamics representation should be reusable across trajectories, controllers, objectives, and platform configurations.

Classical first-principles models satisfy some of these requirements~\cite{loianno2016estimation}, but fall short in practice. They encode the known structure of quadrotor dynamics and are computationally efficient, yet real-world flight behavior is shaped by complex and platform-specific effects that are difficult to model exactly. Small changes such as payload attachment, propeller replacement, motor degradation, or sensor bias can significantly alter the system response. As a result, analytical models often require extensive system identification~\cite{eschmann2024data} and manual tuning, while still struggling to maintain accuracy across operating regimes.

This limitation has motivated a large body of work on neural dynamics models~\cite{saviolo2022physics, bauersfeld2021neurobem, rao2024learning, saviolo2023learning, duong2021hamiltonian, punjani2015deep, hewing2019cautious}. These methods aim to learn unmodeled nonlinear effects directly from data and have been explored in both offline and online settings. Offline methods train models from pre-collected trajectory datasets~\cite{saviolo2022physics, bauersfeld2021neurobem, rao2024learning, duong2021hamiltonian, punjani2015deep, hewing2019cautious}, while online methods adapt the model during deployment to account for changing dynamics or distribution shift~\cite{fu2016one, saviolo2023active, wang2018safe, lew2022safe, o2022neural, jiahao2023online}. Despite these differences, most approaches share a common formulation: they learn a predictive encoder--decoder model that estimates the next state or observation and recursively feeds this prediction back as input during rollout.

While intuitive, this autoregressive predictive formulation violates a central requirement for model-based control: long-horizon stability. Since the model is repeatedly conditioned on its own imperfect outputs, small one-step errors arising from uncertainty, approximation, or inaccuracies can accumulate over time, leading to drift, instability, and physically implausible trajectories. In practice, the error may easily compound and  can severely degrade rollout fidelity as the prediction horizon increases~\cite{rao2024learning}. Adding physics-inspired regularization~\cite{saviolo2022physics}, residual structures~\cite{bauersfeld2021neurobem}, or online adaptation~\cite{saviolo2023active} can improve local prediction accuracy, but does not remove the underlying autoregressive error accumulation mechanism. Moreover, reconstructing the full future state or observation at every step can force the representation to preserve nuisance factors such as sensor noise, bias, or task-irrelevant variation. This becomes especially problematic for high-dimensional observations, where full observation prediction is computationally expensive.

Model-Based Reinforcement Learning (MBRL) provides an alternative route by jointly learning a world model and a task-driven policy from interaction data~\cite{robine2023transformerbased, finn2017deep, wu2021exampledriven}. Although such approaches can achieve strong task performance, the learned model is often coupled to a particular reward, policy, or visitation distribution. Consequently, the resulting representation hardly generalizes across controllers, reference trajectories, or downstream tasks. For aerial robotics, where the same dynamics model should ideally support different tasks, such task-coupled learning can be restrictive.

A further obstacle is data collection. For any learned dynamics model, the quality of the dataset is as important as the choice of architecture. The data collection process must informative (e.g., efficiently covering the overall flight envelope) and sample efficient and therefore covering a diverse range of states, control inputs, speeds, accelerations, and operating conditions. However, for aerial robots, there is currently no systematic procedure for collecting such a dataset. In practice, data collection often relies on hand-designed trajectories, expert pilots, or task-specific flight logs. This makes the resulting dataset incomplete and biased toward a limited set of behaviors. Collecting broader real-world data is also expensive, risky, and labor-intensive. It often requires controlled environments and aggressive maneuvers, which increase the likelihood of crashes or hardware damage. Moreover, when the platform configuration changes, existing pipelines often require new data collection and retraining, limiting scalability.

These limitations raise the central question of this work: \emph{Can we learn an efficient dynamic model for aerial robots that provides accurate long-horizon prediction, physically interpretable rollouts, real-time control capability, and zero-shot generalization across tasks, while reducing dependence on risky real-world data collection?} To address this question, we propose a domain-randomized simulation-to-real framework for training a JEPA-style latent dynamics model for real-time aerial control. Rather than reconstructing future states or observations autoregressively, our model learns to predict future representations in a structured latent space. This formulation encourages the model to capture control-relevant dynamics while avoiding unnecessary reconstruction of nuisance details. To make the learned representation usable for model-based control, we introduce a physics-inspired prober that maps frozen latent rollouts to physically meaningful state variables through a lightweight kinematic structure. By combining latent predictive representation learning, physical structure, and domain-randomized simulation, our approach enables accurate long-horizon prediction and real-time control while remaining reusable across tasks and platform variations. Our main contributions are summarized as follows:

\begin{itemize}
    \item We introduce a JEPA-styled latent dynamics model with a physics-inspired prober that maps frozen representations to physically meaningful state variables, enabling physically grounded long-horizon prediction for quadrotor control.
    \item We demonstrate that by integrating the learned latent dynamics model within a sampling-based optimization framework, we exploit the predictive capabilities of the learned model for real-time control and robust zero-shot sim-to-real transfer on resource-constrained embedded platforms.
    \item We propose a domain-randomized simulation pipeline for automated dataset generation, reducing the need for extensive and potentially unsafe real-world data collection.
    \item We extensively evaluate the proposed framework against current dynamics learning baselines and key design choices through open-loop prediction, and outdoor closed-loop control experiments, demonstrating improved long-horizon accuracy, robust control, and generalization across trajectories and platform variations.
\end{itemize}

\section{Related Works}
\subsection{Predictive Modeling}
A large body of work has studied learning robot dynamics from data in both offline~\cite{punjani2015deep, brunton2016discovering, duong2021hamiltonian} and online settings. Prior offline approaches have explored residual learning over nominal models~\cite{bauersfeld2021neurobem, kulathunga2024residual}, improved multi-step training objectives~\cite{rao2024learning}, and physics-inspired regularization~\cite{saviolo2022physics}.  On the other hand, online methods~\cite{fu2016one, saviolo2023active, wang2018safe, lew2022safe, o2022neural, jiahao2023online} adapt the model during deployment, typically by finetuning the last few layers of the neural dynamics model. Although these directions improve the ability to capture dynamics, each has an inherent limitation. Residual learning strongly depends on the fidelity of the nominal model. Physics-inspired regularization can guide training toward more plausible solutions, but it does not guarantee physically meaningful rollouts. Online adaptation adds deployment complexity because it requires continual on-robot data collection and model updates, along with extra compute and safeguards to prevent unstable intermediate models from degrading control. Moreover, all these approaches remain based on an autoregressive encoder--decoder paradigm, in which predicted states or observations are recursively reused as inputs for future prediction~\cite{bansal2016learning, mohajerin2019multistep, looper2022temporal, saviolo2022physics}. As a result, small one-step errors are propagated forward and accumulate over time, leading to drift and degraded long-horizon fidelity. This compounding-error problem is intrinsic to predictive modeling and is not removed by better losses, residual corrections, or online updates alone, motivating the need for alternatives. In contrast, the proposed JEPA-styled approach produces more accurate long-horizon model predictions compared to these classic predictive models by properly capturing the temporal and spatial context of the problem using a compact abstract representation, avoiding reconstruction altogether.

\subsection{Joint Embedding Predictive Architectures}
JEPA-style methods offer an appealing alternative to predictive encoder--decoder models by learning to predict future embeddings rather than reconstructing full future observations, yielding compact representations that are naturally scalable to high-dimensional inputs. This idea has shown promise in image representation learning~\cite{assran2023self}, video prediction~\cite{bardes2023v, assran2025v, bardes2023mc}, and more recently in robotics~\cite{zhou2024dino}. However, most JEPA-style robotics works have focused on vision-centric settings such as manipulation~\cite{goswami2025world} and navigation~\cite{bar2025navigation}, with several demonstrated only on toy problems or in exocentric simulation without real-world validation~\cite{yin2026ddp, maes2026leworldmodel, sobal2025learning}. Moreover, these methods are typically studied at the level of high-level kinematic planning, where abstract latent predictions are often sufficient. Their use for feedback control remains largely unexplored, particularly for aerial robots, where high-frequency inference of the system dynamics and real-time compute constraints make control-oriented latent modeling significantly more challenging. This is what we address in this work by proposing by proposing the first JEPA framework trained end-to-end for real-time quadrotor control.

\subsection{Model-Based Reinforcement Learning}
With the rise of deep Reinforcement Learning (DRL), a new class of controllers has emerged that leverage learned models of system dynamics to plan and act~\cite{lambert2019low, chua2018deep, williams2017information, robine2023transformerbased, li2025robotic}. These approaches aim to improve data efficiency and generalization by explicitly reasoning about future system evolution, rather than relying solely on reactive, model-free policies. Early model-based RL methods primarily focused on learning dynamics in state space~\cite{hafner2019learning}, and have since been extended to handle high-dimensional sensory inputs such as images~\cite{micheli2022transformers}. However, predicting directly in observation space is often data-intensive and computationally expensive, particularly for vision-based inputs, and can lead to the learning of noisy features when the training data distribution is imperfect. Latent-space prediction offers a more compact alternative, but many existing approaches rely on reconstruction-based objectives~\cite{hafner2025mastering}, inheriting similar limitations associated with observation reconstruction. Moreover, a large class of model-based RL methods incorporate reward prediction, either explicitly or as an auxiliary objective, when learning latent representations~\cite{verraest2025skydreamer, krinner2025accelerating, romero2025dream, nan2025efficient}, which inherently couples the learned world model to a specific task. In contrast, in this work we, decouple task-dependent information from latent dynamics prediction to enable general-purpose, real-time quadrotor control without any task-specific reward conditioning.

\section{Background} 
\label{sec:background}

\subsection{Learning System Dynamics}

We consider a discrete-time dynamical system with state 
$\mathbf{x}_{t} \in \mathbb{R}^{D_s}$ and control action 
$\mathbf{a}_{t} \in \mathbb{R}^{D_a}$. 
The dynamics are described by a transition function 
$f : \mathbb{R}^{D_s} \times \mathbb{R}^{D_a} \rightarrow \mathbb{R}^{D_s}$
\begin{equation}
    \mathbf{x}_{t+1} = f(\mathbf{x}_{t}, \mathbf{a}_{t}) .
\end{equation}
For a quadrotor, we define the state as
\begin{equation}
    \mathbf{x}_t =
    \begin{bmatrix}
        \mathbf{p}_t^\top &
        \mathbf{v}_t^\top &
        \mathbf{r}_{x,t}^\top &
        \mathbf{r}_{y,t}^\top &
        \mathbf{r}_{z,t}^\top &
        \boldsymbol{\omega}_t^\top
    \end{bmatrix}^\top .
\end{equation}

Here, $\mathbf{p}_{t} \in \mathbb{R}^{3}$ and 
$\mathbf{v}_{t} \in \mathbb{R}^{3}$ denote position and velocity in the inertial frame. 
The attitude is represented by the rotation matrix 
$\mathbf{R}_t =
\begin{bmatrix}
    \mathbf{r}_{x,t} &
    \mathbf{r}_{y,t} &
    \mathbf{r}_{z,t}
\end{bmatrix}
\in \mathrm{SO}(3)$, whose columns are 
$\mathbf{r}_{x,t}, \mathbf{r}_{y,t}, \mathbf{r}_{z,t} \in \mathbb{R}^{3}$. This rotation matrix represents the attitude of the body frame in the inertial frame. The angular velocity $\boldsymbol{\omega}_{t} \in \mathbb{R}^{3}$ is expressed in the body frame. The action $\mathbf{a}_{t} = [f_{0,t}, f_{1,t}, f_{2,t}, f_{3,t}]^\top \in \mathbb{R}^{4}$ denotes the four motor forces.

\subsection{Predictive Modeling}
\label{sec::predictive_modelling}
A common approach to learning dynamics is to train a predictive model on state-action transitions. 
Let 
$\mathcal{D} =
\{(\mathbf{x}_{i}, \mathbf{a}_{i}, \mathbf{x}_{i+1})\}_{i=1}^{N}$
be a dataset of $N$ transitions. The goal is to approximate the dynamics using a neural network $ h$, parameterized by weights $\boldsymbol{\theta}$. Formally, the state at the next time index, $i + 1$ is given by
\begin{equation}
    \tilde{\mathbf{x}}_{i+1} = h_{\boldsymbol{\theta}}(\mathbf{x}_{i}, \mathbf{a}_{i}).
\end{equation}
 
\noindent The model is trained by minimizing the prediction loss
\begin{equation}
    \min_{\boldsymbol{\theta}} 
    \frac{1}{N} 
    \sum_{i=1}^{N}
    \left\|
        \mathbf{x}_{i+1}
        -
        h_{\boldsymbol{\theta}}(\mathbf{x}_{i}, \mathbf{a}_{i})
    \right\|_2^2 .
\label{eq:one_step_loss}
\end{equation}
This objective encourages accurate local predictions. However, model-based control requires prediction over a sequence of future actions. Given an action sequence, the model is rolled out recursively
\begin{equation}
    \hat{\mathbf{x}}_{t+T} =
    h_{\boldsymbol{\theta}}
    ( \dots
        h_{\boldsymbol{\theta}}
        (
            h_{\boldsymbol{\theta}}(\mathbf{x}_t, \mathbf{a}_t),
        \mathbf{a}_{t+1}
        ) 
    \dots , \mathbf{a}_{t+T}
    )
    .
\end{equation}

This recursive formulation introduces compounding error. Let the vector prediction error be $\boldsymbol{\epsilon}_{t+k} = \mathbf{x}_{t+k} - \tilde{\mathbf{x}}_{t+k}$. During rollout, each prediction is used as the input to the next step,
\begin{equation}
    \begin{aligned}
        \tilde{\mathbf{x}}_{t+1} &= 
        h_{\boldsymbol{\theta}}(\mathbf{x}_t, \mathbf{a}_t),\\
        \tilde{\mathbf{x}}_{t+2} &= 
        h_{\boldsymbol{\theta}}(\mathbf{x}_{t+1} - \boldsymbol{\epsilon}_{t+1}, \mathbf{a}_{t+1}),\\
        &\vdots\\
        \tilde{\mathbf{x}}_{t+T} &= 
        h_{\boldsymbol{\theta}}(\mathbf{x}_{t+T-1} - \boldsymbol{\epsilon}_{t+T-1}, \mathbf{a}_{t+T-1}).
    \end{aligned}
\label{eq:compounding_error}
\end{equation}

\noindent This makes the source of compounding explicit. At each step, the model is evaluated on a state that is shifted from the true trajectory by the accumulated prediction error. As the horizon increases, these errors alter the future inputs to the model. Thus, small one-step errors can grow into large rollout errors. This can lead to drift and physically implausible trajectories.

\begin{figure*}[t]
    \vspace*{4pt} 
    \centering
    \includegraphics[width=\textwidth, trim=0 335 0 35, clip]{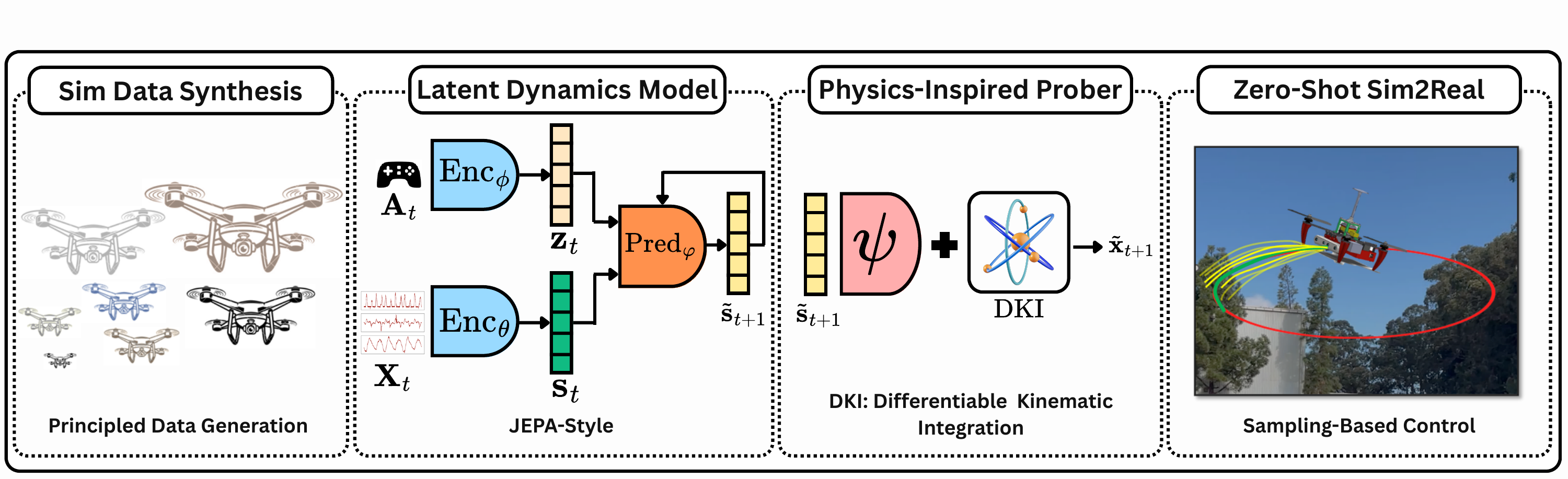}
    \caption{Overview of the proposed framework. We learn a JEPA-style latent dynamics model with a physics-inspired prober that maps abstract embeddings to physically meaningful states, enabling stable long-horizon quadrotor prediction. Trained entirely on domain-randomized simulation data and deployed within a sampling-based controller, the model achieves real-time execution and robust zero-shot sim-to-real transfer validated in outdoor flight experiments.}
    \label{fig:methodology}
    \vspace{-10pt}
\end{figure*}

\section{Methodology}
\label{sec:methodology}
In this section, we describe the methodology underlying our approach (see Figure~\ref{fig:methodology}). We first introduce our proposed JEPA-styled dynamics learning formulation. Next, we present the training objective for learning dynamics. We then describe a sampling-based optimization framework that leverages the learned model for optimal real-time quadrotor control. Lastly, we describe automated data collection framework, including trajectory generation, closed-loop control, and domain randomization strategy used to generate diverse trajectory-level data.

\subsection{Problem Formulation}
\label{sec:latent_dynamics}

We seek to learn a dynamics model that supports long-horizon prediction for model-based quadrotor control. As discussed in Section~\ref{sec::predictive_modelling}, directly predicting future states in an autoregressive manner can lead to compounding errors. Therefore, we formulate dynamics learning in a compact latent space. This follows the JEPA principle of predicting representations rather than reconstructing inputs~\cite{lecun2022path}. The goal is to learn a representation that captures the evolution of the system while avoiding unnecessary reconstruction of task-irrelevant details. At each time step, the UAV receives an estimated full-state observation $\mathbf{x}_t$. To provide temporal context and promote better sim-to-real transfer, we condition the model on histories of states and actions over a window of length $H$. The state history and action history are defined as
\begin{equation}
    \mathbf{X}_t =
    \begin{bmatrix}
        \mathbf{x}_{t-H}^{\top} &
        \cdots &
        \mathbf{x}_{t}^{\top}
    \end{bmatrix}^{\top},
    \qquad
    \mathbf{A}_t =
    \begin{bmatrix}
        \mathbf{a}_{t-H}^{\top} &
        \cdots &
        \mathbf{a}_{t}^{\top}
    \end{bmatrix}^{\top}.
\end{equation}

\noindent This history-based formulation gives the model access to recent motion and actuation trends.  It also helps account for effects that are not fully captured by a single state, such as actuator delay, drag, sensor noise, and platform-dependent dynamics. The state and action histories are encoded into latent representations $\mathbf{s}_t = \text{Enc}_\theta(\mathbf{X}_t)$ and $\mathbf{z}_t = \text{Enc}_\phi(\mathbf{A}_t)$. The latent dynamics predictor then estimates the next latent state as $\tilde{\mathbf{s}}_{t+1} = \text{Pred}_\varphi(\mathbf{s}_t,\mathbf{z}_t)$. For a prediction horizon of $T$, the predictor is recursively unrolled using the encoded actions 
\begin{equation}
\tilde{\mathbf{s}}_{t+T} =
\text{Pred}_\varphi(
\dots
\text{Pred}_\varphi(
\text{Pred}_\varphi(\mathbf{s}_t,\mathbf{z}_t),
\mathbf{z}_{t+1}),
\dots,
\mathbf{z}_{t+T-1}).
\end{equation}
Thus, the learning problem is to train the encoders and predictor so that the latent rollout 
$\{\tilde{\mathbf{s}}_{t+1}, \dots, \tilde{\mathbf{s}}_{t+T}\}$ remains consistent with the encoded future trajectory 
$\{\mathbf{s}_{t+1}, \dots, \mathbf{s}_{t+T}\}$. 
This formulation avoids direct reconstruction of future states while preserving the information needed for long-horizon dynamics.

\subsection{Training Objective}
\label{sec:training_objective}

\begin{figure*}[t]
    \vspace*{4pt} 
    \centering
    \includegraphics[width=\textwidth, trim=0 440 0 0, clip]{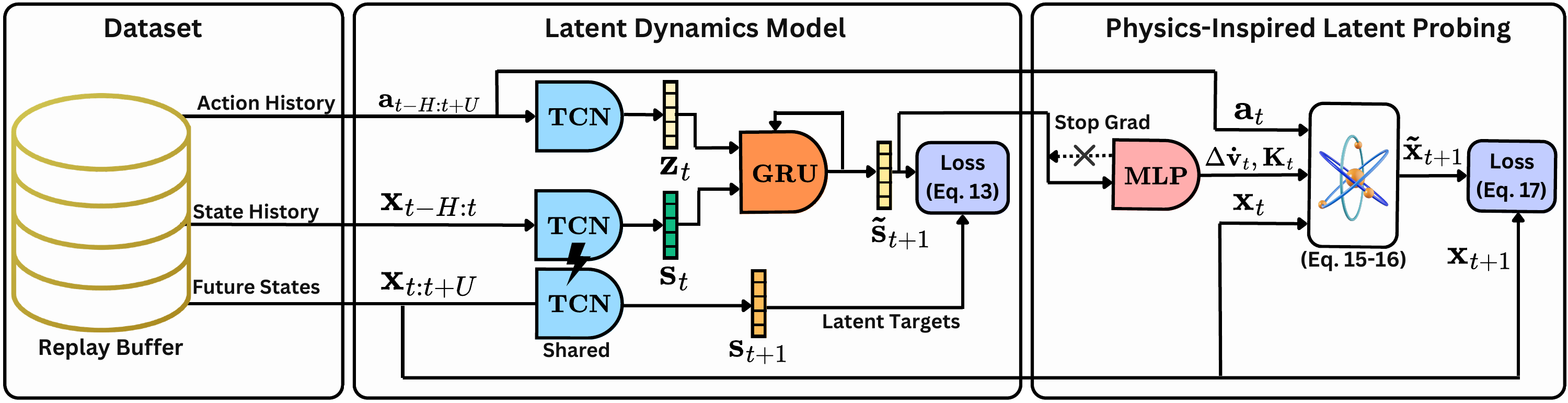}
    \caption{Two-stage training pipeline of our proposed approach. In the first stage, an encoder maps a history of past states into a latent representation, and a predictor propagates this latent forward in time conditioned on a sequence of control actions; in the second stage, a physics-inspired prober is trained on frozen latent embeddings to map them to physically grounded states.}
    \label{fig:dynamics_learning_framework}
    \vspace{-10pt}
\end{figure*}

The objective is to learn latent representations that support accurate multi-step dynamics prediction for long-horizon forecasting (see Figure~\ref{fig:dynamics_learning_framework}). The proposed loss contains two terms. The first term enforces predictive consistency between rolled-out latent predictions and encoded future states. The second term regularizes the latent space to prevent representation collapse. We define the multi-step latent prediction loss as
\begin{equation}
\mathcal{L}_{\mathrm{pred}}
=
\frac{1}{T}
\sum_{k=1}^{T}
\|
\tilde{\mathbf{s}}_{t+k}
-
\mathbf{s}_{t+k}
\|_2^2.
\end{equation}
This loss encourages the encoder and predictor to learn representations that remain predictive over the full rollout horizon. However, minimizing this term alone admits degenerate solutions. For example, the encoder could map all inputs to nearly constant embeddings, yielding low prediction error without preserving meaningful system dynamics. We therefore utilize an anti-collapse regularization term. We employ Sketched Isotropic Gaussian Regularization (SIGReg)~\cite{balestriero2025lejepa}, which encourages the latent embeddings to match an isotropic Gaussian distribution. This promotes diversity and isotropy in the representation space.  

To apply SIGReg over a rollout during training, we collect the predicted latent embeddings into a tensor $\mathbf{S} \in \mathbb{R}^{T \times B \times D}$, where $B$ is the training batch size during training, and $D$ is the embedding dimension.
Instead of matching the full $D$-dimensional latent distribution directly, SIGReg compares random one-dimensional projections of the latent distribution to a standard Gaussian. 
We sample $M$ random unit vectors, where the directions are sampled uniformly on the hypersphere,
$\{\boldsymbol{\xi}_{m}\}_{m=1}^{M} \subset \mathbb{S}^{D-1}$ 
and project the latent tensor along each direction $\mathbf{h}^{(m)} = \mathbf{S}\boldsymbol{\xi}_{m} \in \mathbb{R}^{T \times B} $. For each projection, we evaluate the univariate Epps--Pulley test statistic, measuring the distribution mismatch. Let
\begin{equation}
\phi_N(t;\mathbf{h}^{(m)}) = \frac{1}{B}\sum_{b=1}^{B} e^{it h_b^{(m)}},
\end{equation}
denote the empirical characteristic function of the projected samples, and let $\phi_0(t)$ denote the characteristic function of the standard Gaussian $\mathcal{N}(0,1)$. 
The projected discrepancy is
\begin{equation}
T^{(m)}
=
\int_{-\infty}^{\infty}
w(t)
\left|
\phi_B(t;\mathbf{h}^{(m)}) - \phi_0(t)
\right|^2 dt,
\end{equation}
where $w(t)$ is a weighting function, typically chosen as a Gaussian kernel. 
The SIGReg penalty is then
\begin{equation}
\mathcal{L}_{\mathrm{SIGReg}}
=
\frac{1}{M}\sum_{m=1}^{M} T^{(m)}.
\end{equation}
By the Cram\'er--Wold theorem, matching all one-dimensional marginals is equivalent to matching the full joint distribution. 
Therefore, minimizing these projected discrepancies encourages the latent distribution to approach an isotropic Gaussian. In practice, the integral in $T^{(m)}$ is evaluated numerically using quadrature, following~\cite{balestriero2025lejepa}. We average this SIGReg penalty across the temporal dimension. The final objective combines multi-step prediction consistency with SIGReg regularization
\begin{equation}
\mathcal{L}_{\mathrm{total}}
=
\mathcal{L}_{\mathrm{pred}}
+
\lambda_{\mathrm{sig}}\mathcal{L}_{\mathrm{SIGReg}}.
\label{eq::sigreg}
\end{equation}

\noindent Here, $\lambda_{\mathrm{sig}}$ controls the strength of the anti-collapse regularization relative to the prediction objective. The SIGReg component introduces only two practical hyperparameters, $M$ and $\lambda_{\mathrm{sig}}$, with $\lambda_{\mathrm{sig}}$ being the main parameter to tune. Following~\cite{balestriero2025lejepa}, we also find that performance is not highly sensitive to the number of random projections $M$. This is in contrast to many self-supervised representation learning objectives that require balancing several regularization terms~\cite{bardes2021vicreg, sobal2025learning}, stop-gradient~\cite{grill2020bootstrap} design choices, exponential moving averages, or reconstruction weights.

\subsection{Physics-Inspired Probing Mechanism}
\label{sec:physics_prober}
For model-based control, latent predictions must be converted into physically meaningful quantities. This is necessary for evaluating tracking costs, enforcing state constraints, and respecting actuator limits. However, the JEPA-style dynamics model predicts future evolution in an abstract representation space. We therefore introduce a physics-inspired probing mechanism that maps latent rollouts to interpretable state trajectories through a differentiable kinematic model. After training the latent dynamics model with eq.~(\ref{eq::sigreg}), we perform a second training stage for metric-state recovery. During this stage, the encoders and predictor $(\mathrm{Enc}_\theta, \mathrm{Enc}_\phi, \mathrm{Pred}_\varphi)$, are frozen and only the probing network is optimized. This separation prevents the supervised state-recovery objective from altering the learned latent dynamics, while allowing the prober to learn a physically grounded map from frozen latent rollouts to metric state trajectories.

Given the predicted latent sequence $\{\tilde{\mathbf{s}}_{t+1}, \dots, \tilde{\mathbf{s}}_{t+T}\}$, the current estimated state $\mathbf{x}_t$, and the control sequence $\{\mathbf{a}_{t}, \dots, \mathbf{a}_{t+T-1}\}$ used for latent rollout which in our specific case will be generated using the optimal control procedure presented in Section~\ref{sec:MPPI}. The probing network $\psi$ predicts residual correction terms
\begin{equation}
    \{\Delta \mathbf{\dot{v}}_{t+k}, \mathbf{K}_{t+k}\}
    =
    \psi(\tilde{\mathbf{s}}_{t+k}).
\label{eq::prober}
\end{equation}
Here, $\Delta \mathbf{\dot{v}}_{t+k} \in \mathbb{R}^3$ represents a residual translational acceleration, $\mathbf{K}_{t+k} \in \mathbb{R}^{3\times4}$ parameterizes residual angular acceleration, and $k$ is the unroll index. The control input is the individual rotor forces as mentioned in Section~\ref{sec:background}. The predicted state trajectory is obtained by integrating a residual-corrected kinematic model. At each step, the translational acceleration is given by the nominal thrust-induced acceleration plus the learned residual
\begin{equation}
\begin{aligned}
    \mathbf{\dot{v}}_{t}
    &=
    \frac{\sum_{i=0}^{3} f_{i,t}}{m}
    \mathbf{R}_{t}\mathbf{e}_3
    -
    \mathbf{g}
    +
    \Delta \mathbf{\dot{v}}_{t}, \\
    \Delta \boldsymbol{\tau}_{t}
    &=
    \mathbf{K}_{t}\mathbf{a}_{t}.
\end{aligned}
\label{eq:prober_residuals}
\end{equation}
Here, $\mathbf{e}_3=[0~0~1]^\top$, and $\Delta \boldsymbol{\tau}_{t+k}$ denotes the latent-conditioned residual angular acceleration. The state is then propagated using
\begin{equation}
\begin{aligned}
    \mathbf{p}_{t+1}
    &=
    \mathbf{p}_{t}
    +
    \mathbf{v}_{t}\Delta t, \\
    \mathbf{v}_{t+1}
    &=
    \mathbf{v}_{t}
    +
    \mathbf{\dot{v}}_{t}\Delta t, \\
    \mathbf{R}_{t+1}
    &=
    \mathbf{R}_{t}
    \exp\!\left(
        [\boldsymbol{\omega}_{t}]_\times \Delta t
    \right), \\
    \boldsymbol{\omega}_{t+1}
    &=
    \boldsymbol{\omega}_{t}
    +
    \Delta \boldsymbol{\tau}_{t}\Delta t .
\end{aligned}
\label{eq:physics_prober_integrator}
\end{equation}
This compact integrator preserves the geometric structure of the attitude dynamics through the $\mathrm{SO}(3)$ exponential map, while allowing the latent representation to correct for unmodeled translational and rotational dynamics effects.

During prober training, a stop-gradient operation is applied to the predicted latent embeddings before they are passed to $\psi$. This prevents the supervised state-recovery loss from modifying the learned latent dynamics. Let $\tilde{\mathbf{x}}_{t+k}$ denote the state obtained by integrating eq.~\eqref{eq:physics_prober_integrator}. The probing network is optimized using the supervised rollout loss.
\begin{equation}
\mathcal{L}_{\text{prober}}
=
\frac{1}{T}
\sum_{k=1}^{T}
\|
\tilde{\mathbf{x}}_{t+k}
-
\mathbf{x}_{t+k}
\|_2^2 .
\end{equation}
By combining frozen latent predictions with a structured differentiable integrator, the probing mechanism converts abstract representations into physically meaningful rollouts suitable for control.

\subsection{Sampling-Based Control with Learned Dynamics}~\label{sec:MPPI}

We consider the problem of tracking a reference trajectory over a finite horizon $T$. 
At each control timestep $t$, given the current estimated state history $\mathbf{X}_t$ and action history $\mathbf{A}_t$, the objective is to compute a sequence of future control inputs 
$\{\mathbf{a}_{t}, \dots, \mathbf{a}_{t+T-1}\}$ 
that minimizes a trajectory tracking cost with respect to desired reference states 
$\{\mathbf{x}^{\mathrm{ref}}_{t+1}, \dots, \mathbf{x}^{\mathrm{ref}}_{t+T}\}$.

We integrate our learned dynamics model within a sampling-based optimization framework, MPPI, where future action sequences are optimized using Monte Carlo sampling.

\textbf{Action Sampling}.
Let the current nominal action sequence be 
$\mathbf{a}^{\mathrm{nom}}=\{\mathbf{a}^{\mathrm{nom}}_{0}, \dots, \mathbf{a}^{\mathrm{nom}}_{T-1}\}$.
We generate $S$ candidate sequences by perturbing it
\begin{equation}
    \mathbf{a}^{(s)}_{k}
    =
    \mathbf{a}^{\mathrm{nom}}_{k}
    +
    \boldsymbol{\epsilon}^{(s)}_{k},
    \qquad
    \boldsymbol{\epsilon}^{(s)}_{k}
    \sim
    \mathcal{N}(0,\boldsymbol{\Sigma}),
\end{equation}
for $s=1,\dots,S$ and $k=0,\dots,T-1$, with covariance matrix $\boldsymbol{\Sigma}$. The admissible action set is defined element-wise as
\begin{equation}
\mathcal{A}
=
\left\{
\mathbf{a} \in \mathbb{R}^{4}
\;\middle|\;
\mathbf{a}_{\min}
\le
\mathbf{a}
\le
\mathbf{a}_{\max}
\right\},
\end{equation}
where $\mathbf{a}_{\min}$ and $\mathbf{a}_{\max}$ denote lower and upper limits. Each sampled action is projected onto $\mathcal{A}$ via element-wise clamping
\begin{equation}
\mathbf{a}^{(s)}_{k}
\leftarrow
\Pi_{\mathcal{A}}\!\left(\mathbf{a}^{(s)}_{k}\right),
\end{equation}
where $\Pi_{\mathcal{A}}(\cdot)$ denotes element-wise clamping.

\textbf{Latent Rollout and State Prediction}. 
For each sampled action sequence, we append these actions to the action history previously executed to construct the updated action sequence $\{\mathbf{a}^{(s)}_{t-H}, \dots, \mathbf{a}^{(s)}_{t}, \dots, \mathbf{a}^{(s)}_{t+T-1}\}$. The context states and actions are encoded using $\text{Enc}_\theta$ and $\text{Enc}_\phi$, respectively. For each rollout sample $s$, the latent dynamics are recursively unrolled using $\mathrm{Pred}_\varphi$. The predicted latent sequence 
$\{\tilde{\mathbf{s}}^{(s)}_{t+1},\dots,\tilde{\mathbf{s}}^{(s)}_{t+T}\}$
is mapped to physically grounded states via the probing mechanism, eqs.~(\ref{eq::prober})--(\ref{eq:physics_prober_integrator}), yielding predicted physical states $\{\tilde{\mathbf{x}}^{(s)}_{t+1}, \dots, \tilde{\mathbf{x}}^{(s)}_{t+T}\}$.

\textbf{Trajectory Cost}.
Each sampled trajectory is evaluated using a running cost of the form
\begin{equation}
\mathcal{J}^{(s)}
=
\frac{1}{T}
\sum_{k=1}^{T}
\ell\!\left(
\tilde{\mathbf{x}}^{(s)}_{t+k},
\mathbf{x}^{\mathrm{ref}}_{t+k},
\mathbf{a}^{(s)}_{t+k-1}
\right),
\end{equation}
where the per-step cost consists of weighted quadratic tracking errors and control effort:
\begin{equation}
\begin{aligned}
\ell(.)
&=
\|\tilde{\mathbf{x}}^{(s)}-\mathbf{x}^{\mathrm{ref}}\|_{\mathbf{Q}_{\text{x}}}^2
+
\|\mathbf{a}^{(s)} - \mathbf{a}^{\mathrm{ref}}\|_{\mathbf{Q}_{\text{a}}}^2,
\end{aligned}
\end{equation}
where $\mathbf{Q}_{\text{x}}$ and $\mathbf{Q}_{\text{a}}$ are constant positive diagonal weight matrices for state and control, respectively.

\textbf{Softmax Weighting and Control Update}.
Following the MPPI formulation~\cite{williams2017information}, we compute importance weights
\begin{equation}
w^{(s)}
=
\frac{
\exp\!\left(
-\frac{1}{\lambda}
(\mathcal{J}^{(s)} - \mathcal{J}_{\min})
\right)
}{
\sum_{r=1}^{S}
\exp\!\left(
-\frac{1}{\lambda}
(\mathcal{J}^{(r)} - \mathcal{J}_{\min})
\right)
},
\end{equation}
where $\mathcal{J}_{\min} = \min_s \mathcal{J}^{(s)}$ and $\lambda>0$ is the temperature parameter. The nominal sequence is then updated as
\begin{equation}
\mathbf{a}^{\mathrm{nom}}_{k}
\leftarrow
\mathbf{a}^{\mathrm{nom}}_{k}
+
\sum_{s=1}^{S}
w^{(s)}\,\delta \mathbf{a}^{(s)}_{k},
\qquad
\delta \mathbf{a}^{(s)}_{k} = \mathbf{a}^{(s)}_{k} - \mathbf{a}^{\mathrm{nom}}_{k}.
\label{eq:mppi_update}
\end{equation}

\noindent Only the first action $\mathbf{a}^{\mathrm{nom}}_{0}$ is executed, and the procedure is repeated in a receding-horizon fashion.

\subsection{Automated Data Synthesis}
\label{sec:data_collection}

The performance of a learned dynamics model is strongly determined by the data distribution used for training. In our case, we show a princpled procedure to synthetize data in simuation to effectively learn zero-shot sim2real transfer. A useful dataset should satisfy three requirements. First, it should be \emph{diverse} to properly represent the overall flight envelope: the trajectories must cover a broad range of positions, velocities, accelerations, attitudes, angular velocities, and control inputs. Second, it should be \emph{dynamically feasible}: the recorded transitions should arise from closed-loop execution of physically valid quadrotor dynamics, rather than from arbitrary state sampling. Third, it should be \emph{robust}: the data should include variations in physical parameters so that the learned model does not overfit to a single nominal platform.

\textbf{Reference Trajectory Generation}. The first step in our data collection pipeline is to generate reference trajectories that excite diverse regions of the quadrotor state space. Manually designed trajectories, such as circles, figure-eights, or straight-line paths, cover only a narrow set of flight behaviors. They can also bias the learned model toward a small number of hand-selected motion patterns. To avoid this, we use Gaussian processes to automatically generate randomized, smooth, and diverse reference trajectories. For each trajectory, we sample the desired position independently along each spatial axis from a Gaussian process prior. A Gaussian process, denoted by $\mathcal{GP}(0,k_j)$, defines a distribution over smooth functions with zero mean and covariance kernel $k_j(t,t')$. We write this as,
\begin{equation}
    p_j(t) \sim \mathcal{GP}(0,k_j(t,t')),
    \qquad j \in \{x,y,z\}.
\label{eq::gp}
\end{equation}
Each kernel is chosen as a sum of periodic kernels with different characteristic length scales and periods. This allows the sampled trajectories to contain both slow global motion and faster local variations. The generated references induce a broad range of velocities, accelerations, attitudes, and angular velocities during tracking. We compute the corresponding velocity and acceleration references by differentiation, and then use differential flatness~\cite{mellinger2011minimum} to obtain the full quadrotor reference. The resulting reference trajectories are randomized, smooth, and dynamically rich. They provide a systematic alternative to manually specified trajectory families and form the basis for collecting diverse state--action rollouts in simulation.

\textbf{Closed-Loop Trajectory Tracking}. The generated references are not used directly as training data. Instead, each reference is tracked in simulation to produce physically feasible state--action trajectories. At each time step, a tracking controller computes the action from the current state and a local reference horizon $\mathbf{x}_{t:t+T}^{\text{ref}}$, $\mathbf{a}_{t}^{}
    =
    \pi_{\mathrm{track}}
    \left(
        \mathbf{x}_{t},
        \mathbf{x}_{t:t+T}^{\text{ref}}
    \right),$
where $\pi_{\mathrm{track}}$ is implemented using a combination of nominal Nonlinear Model Predictive Control (NMPC) and Model Predictive Path Integral (MPPI)~\cite{williams2017information}. The two controllers expose the dataset to complementary action distributions. NMPC generates smooth, optimized tracking commands around the nominal dynamics, while MPPI produces sampling-based action sequences with broader local variation. This diversity is important for learning a dynamics model that remains accurate under the control inputs encountered during real-time sampling-based optimization. At the same time, collecting data through closed-loop controllers ensures that the recorded trajectories respect actuator limits, dynamic feasibility, and realistic state--action correlations, unlike independent random sampling of states and controls.

\textbf{Domain-Randomized Quadrotor Simulation}. Data collection is performed in simulation using standard quadrotor rigid-body dynamics~\cite{song2021flightmare}, with aerodynamic drag and first-order motor delay included. To improve robustness and sim-to-real transfer, we do not generate data from a single nominal simulator. Instead, each rollout is collected from a randomized quadrotor model using a similar ancestral sampling approach described in \cite{eschmann2026raptor}. This sampling approach ensures that the parameter distribution is physical plausible. This allows the dataset to reflect not only the nominal platform, but also nearby systems that may arise from modeling errors, hardware variation, or changes in operating conditions. We define the simulator parameter set as
\begin{equation}
    \boldsymbol{\eta}
    =
    \left\{
        m,\;
        \mathbf{D},\;
        \mathbf{J},\;
        \alpha,\;
        k_f,\;
        k_{\tau},\;
        l
    \right\},
\label{eq:sim_parameter_set}
\end{equation}
where $m$ is the mass, $\mathbf{D}$ is the drag matrix, $\mathbf{J}$ is the inertia matrix, $\alpha$ is the motor time constant, $k_f$ and $k_{\tau}$ are the thrust and torque coefficients, and $l$ is the arm length. These parameters capture the main sources of mismatch between simulation and the real platform, including payload changes, actuator variation, and inertial differences. For each rollout $r$, we sample a randomized parameter set $\boldsymbol{\eta}_r$ from a bounded uniform distribution around the nominal parameter set $\bar{\boldsymbol{\eta}}$
\begin{equation}
    \boldsymbol{\eta}_{r}
    \sim
    \mathcal{U}
    \left(
        \bar{\boldsymbol{\eta}} - \Delta\boldsymbol{\eta},
        \bar{\boldsymbol{\eta}} + \Delta\boldsymbol{\eta}
    \right),
\label{eq:domain_randomization_sampling}
\end{equation}
where $\Delta\boldsymbol{\eta}$ defines the randomization range for each parameter. Each generated reference trajectory is therefore tracked under a different plausible realization of the quadrotor dynamics. This exposes the learned model to a family of systems rather than a single nominal model. Consequently, the learned representation is encouraged to capture dynamics that remain consistent across variations in mass, drag, actuation, and inertial properties.


\begin{figure*}[t]
    \vspace*{4pt} 
    \centering
    \includegraphics[width=\textwidth, trim=0 130 0 130, clip]{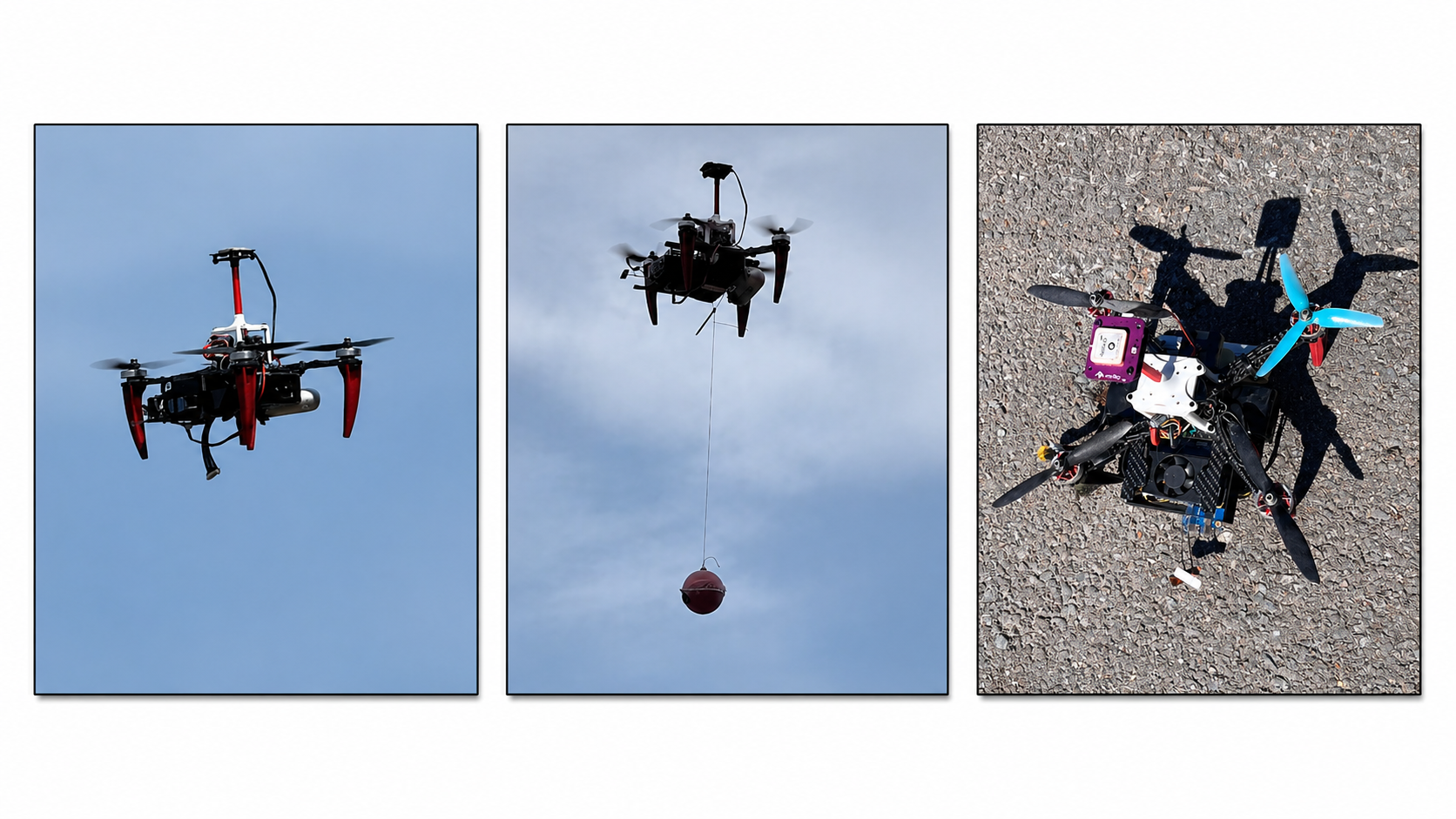}
    \caption{Closed-loop real-world evaluation settings under trajectory tracking. We test the proposed framework under three scenarios: (a) nominal trajectory tracking, (b) payload variation, and (c) propeller switching.}
    \label{fig:Task_definision}
    \vspace{-10pt}
\end{figure*}

\section{Implementation Details}
\subsection{Platform Setup}
Our experiments are conducted on a \(1.3\)-kg quadrotor with a thrust-to-weight ratio of \(4{:}1\). The platform uses an NVIDIA Orin NX for onboard computation and a Pixracer Pro flight controller running PX4~\cite{meier2015px4} for low-level control, with the full software stack integrated through ROS2. A laptop ground station is used only to issue high-level commands such as takeoff, landing, and reference trajectory upload. State estimation is obtained from onboard GPS-based localization, and the we send collective thrust and body-rate commands to the low-level flight controller.  We conducted closed-loop real-world evaluation (see Figure~\ref{fig:Task_definision}) under $3$ trajectory tracking settings: (a) nominal trajectory tracking, (b) payload variation, and (c) propeller switching. All tests are conducted in a larger outdoor flying space $60 \times 70$ m$^2$.

\subsection{Dataset Generation}
To promote sim-to-real transfer, we construct the training dataset through automated trajectory generation and large-scale domain randomization in simulation. For each rollout, the quadrotor parameters are sampled independently from uniform distributions around the nominal platform values, as summarized in Table~\ref{tab:dr_params}. We sample $500$ distinct domains, producing a diverse ensemble of dynamic models that span variations in inertial properties, actuation, motor response, and aerodynamic drag according to the procedure presented in Section~\ref{sec:data_collection}. For each domain, we generate smooth randomized reference trajectories using Gaussian processes~\cite{wiedemann2022training}. The desired position is sampled independently along the $x$, $y$, and $z$ axes using sums of exponential sine-squared periodic kernels (see eq.~(\ref{eq::gp})). Each axis uses three periodic components: the first has length scale $1.3$, while the remaining two have length scales $3.0$ and $4.0$; the periodicities are chosen differently across axes to avoid repetitive motion patterns. Specifically, the $x$ axis uses periodicities $37$, $61$, and $13$, the $y$ axis uses $17$, $23$, and $52$, and the $z$ axis uses $19$, $29$, and $53$. This multi-frequency GP sampling produces trajectories with both slow global motion and faster local variations, which helps excite a broad range of translational and rotational behaviors and properly explore the flight envelope. We generate $20{,}000$ reference trajectories of $10$ s duration, track them in simulation, and record the complete state--action time series. Finally, all trajectories are resampled using cubic splines at a discrete time step of $0.05$ s, corresponding to $20$ Hz, to match the temporal resolution used for multi-step dynamics learning and control. The resulting dataset is split into $80\%$ for training, $10\%$ for validation, and $10\%$ for testing.


\subsection{Network Architecture and Training}
\label{sec::network_architecture}
Both $\mathrm{Enc}_\theta$ and $\mathrm{Enc}_\phi$ are implemented as Temporal Convolutional Networks (TCNs)~\cite{lea2017temporal} with channel sizes $[8,8,16]$ and $[4,4,8]$, respectively. Latent dynamics are modeled using a single-layer GRU predictor~\cite{pmlr-v37-chung15} with hidden dimension $24$ and trained via recursive unrolling over $T=20$ ($1.0$\,s at $20$\,Hz) steps. The history length is set to $H=10$ timesteps ($0.5$\,s at $20$\,Hz). Training is performed for $50$ epochs with batch size $2048$. We employ a SigReg objective with $17$ spline knots and regularization coefficient $\lambda_{\text{sigreg}}=0.02$. Optimization uses Adam with weight decay $10^{-5}$ and gradient clipping at $0.5$. The learning rate follows a linear warmup schedule from $0$ to $5\times10^{-3}$ over $4{,}000$ steps, followed by cosine decay to $1\times10^{-4}$ over $20{,}000$ steps. 

\begin{table}[t]
\centering
\caption{Domain randomization parameters used for simulation data collection.}
\begin{tabular}{l c}
\toprule
\textbf{Parameter} & \textbf{Randomization} \\
\midrule
Mass $m$ (kg) & $\pm 50\%$ nominal \\
Inertia $\mathbf{J}$ (kg\,m$^2$) & $\pm 30\%$ nominal \\
Motor time constant $\alpha$ (s) & $[0.01,\,0.1]$ \\
Drag coefficients $\mathbf{D}$ & $[0.1,\,0.5]$ \\
Thrust coefficient $k_f$ & $\pm 50\%$ nominal \\
Torque coefficient $k_m$ & $\pm 50\%$ nominal \\
\midrule
No. of domains & 500 \\
Total trajectories & 20000 \\
Trajectory duration (s) & 10 \\
\bottomrule
\end{tabular}
\label{tab:dr_params}
\end{table}

\begin{table}[!t]
\centering
\caption{MPPI Parameters}
\begin{tabular}{l c}
\toprule\toprule
\textbf{Parameter} & \textbf{Value} \\
\midrule
Prediction timestep $\Delta t$ & 0.05~s \\
Horizon $T$ & 15 \\
Number of samples $S$ & 512 \\
Temperature $\lambda$ & $10^{-4}$ \\
Action noise $\Sigma$ & $\mathrm{diag}(0.60,0.15,0.15,0.05)$ \\
State cost $\mathbf{Q}$ & $\mathrm{diag}(400,40,20,20)$ \\
Control cost $\mathbf{R}$ & $\mathrm{diag}(0.01,0.05,0.05,0.10)$ \\
\bottomrule\bottomrule
\end{tabular}
\label{tab:mppi_params}
\end{table}

\subsection{MPPI Controller Parameters}
The controller parameters of the proposed framework are summarized in Table~\ref{tab:mppi_params}. To ensure real-time performance, the MPPI controller is implemented entirely in C++. The learned PyTorch latent dynamics model is exported and optimized using NVIDIA TensorRT for accelerated inference on the NVIDIA Jetson Orin NX. This enables high-frequency rollout evaluation and closed-loop control on embedded hardware.

\begin{figure}[t]
  \centering
  \includegraphics[width=1.0\linewidth, trim=0 0 0 0, clip]{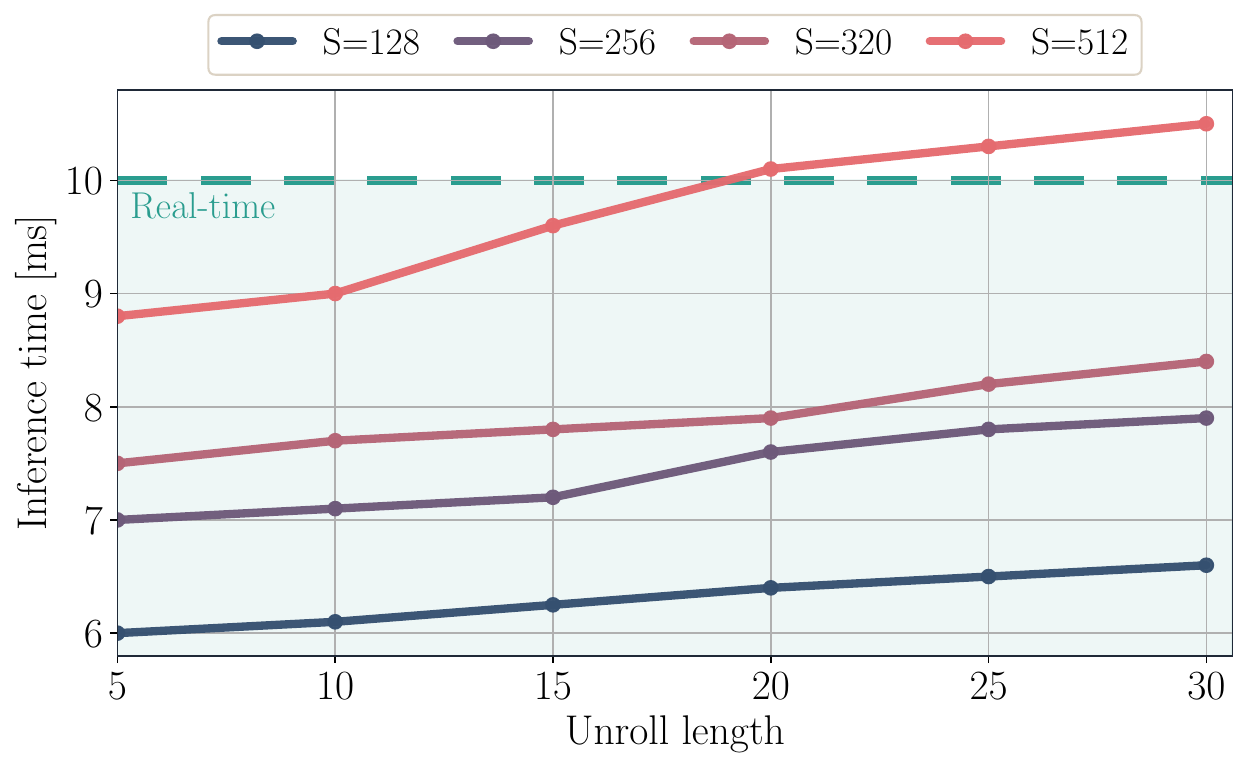}
   \caption{Inference speed on NVIDIA Orin NX. Inference time increases with unroll length and number of MPPI rollouts $S$; the selected horizon is chosen to remain below the 10 ms real-time control budget.}
   \label{fig:inference_speed}
\end{figure}

We analyze the runtime performance of the proposed approach on an NVIDIA Jetson Orin, with all computation executed fully on-board. Figure~\ref{fig:inference_speed} reports the end-to-end control time as a function of rollout horizon \(U\) and sample count \(S\), where each timing includes repeated queries to the learned latent dynamics model within MPPI. Since \(100\) Hz is the minimum rate required for real-time onboard control, the \(10\) ms line defines the feasible operating boundary. As expected, runtime increases with both \(U\) and \(S\), reflecting the higher cost of evaluating more candidate trajectories over longer horizons. The results show that \(U=20\) and \(S=512\) lies near this boundary, making it an effective operating point that maximizes controller lookahead and sample diversity while remaining close to the real-time budget. This trade-off is enabled by the lightweight latent dynamics model described in Section~\ref{sec::network_architecture}, which contains only about \(9\)K parameters and allows the entire framework to run fully on-board on embedded hardware.


\section{Experiments}

We evaluate the proposed world model along two complementary axes: \emph{(i)} offline open-loop prediction, to measure predictive accuracy and long-horizon consistency, and \emph{(ii)} real time onboard closed-loop control on embedded hardware, to assess deployment performance under real-world conditions. Our experimental study is designed to answer the following questions:

\begin{enumerate}
    \item \textbf{Latent Dynamics Modeling:} Does modeling dynamics in a latent space yield more informative and control-relevant representations than direct autoregressive prediction in state space?

    \item \textbf{Physics Interpretability:} Does embedding kinematic structure enable latent rollouts to retain enough information for accurate recovery of metric-state trajectories in open loop?

    \item \textbf{Zero-Shot Sim-to-Real Transfer:} Can a model trained entirely on domain-randomized simulation data transfer to real-world quadrotor navigation without any task-specific fine-tuning?

    \item \textbf{Robustness:} Does the proposed framework remain effective under changes in platform geometry, such as payload attachment or propeller replacement?

    \item \textbf{Data Quality:} How does the quality of the training data distribution affect the predictive accuracy of learned quadrotor dynamics?
\end{enumerate}

\subsection{Latent Dynamics Modeling}


\textbf{Compounding Error Analysis}. We first evaluate whether latent-space dynamics modeling is less susceptible to recursive error accumulation than direct autoregressive prediction in state space. To this end, we use two complementary metrics: the \emph{compounding ratio} (CR) and \emph{error rate} (ER).

The CR compares recursive open-loop rollout against teacher-forced prediction considering the same horizon. Let $e^{\mathbf{x}}_{k,\mathrm{TF}}$ denote the teacher-forced error, where the model is conditioned on the true past states from the dataset, and let $e^{\mathbf{x}}_{k,\mathrm{rollout}}$ denote the open-loop rollout error, where the model is conditioned on its own past predictions. Let as previously defined the state prediction error at time step $k$ as the root-mean-square error between the predicted state $\tilde{\mathbf{x}}_k$ and the ground-truth state $\mathbf{x}_k$
\begin{equation}
e_k^{\mathbf{x}}
=
\sqrt{
\frac{1}{D_{\mathbf{s}}}
\left\|
\tilde{\mathbf{x}}_k - \mathbf{x}_k
\right\|_2^2
},
\label{eq:state_rmse}
\end{equation}
where $D_{\mathbf{s}}$ denotes the dimensionality of the state vector. We then define the compounding ratio as
\begin{equation}
    \mathrm{CR}_k
    =
    \frac{
        e^{\mathbf{x}}_{k,\mathrm{rollout}}
    }{
        e^{\mathbf{x}}_{k,\mathrm{TF}}
    }.
\label{eq::compounding_ratio}
\end{equation}
This ratio isolates the excess error caused by recursive prediction. Values near $1$ indicate that open-loop rollout remains close to teacher-forced performance, while larger values indicate stronger compounding effects. Values below $1$ indicate that the rollout error is lower than the teacher-forced error at that horizon. This suggests that the model's own predicted trajectory is locally easier to predict than the ground-truth trajectory, or that teacher-forced inputs introduce larger local inconsistencies.

The second metric is defined as error growth, the expected increase in error between consecutive recursive predictions
\begin{equation}
    \mathrm{ER}_k
    =
    \mathbb{E}
    \left[
        e_k^{\mathbf{x}}
        -
        e_{k-1}^{\mathbf{x}}
    \right].
\end{equation}

\begin{figure*}[t]
    \vspace*{4pt} 
    \centering
    \includegraphics[width=\textwidth, trim=0 0 0 0, clip]{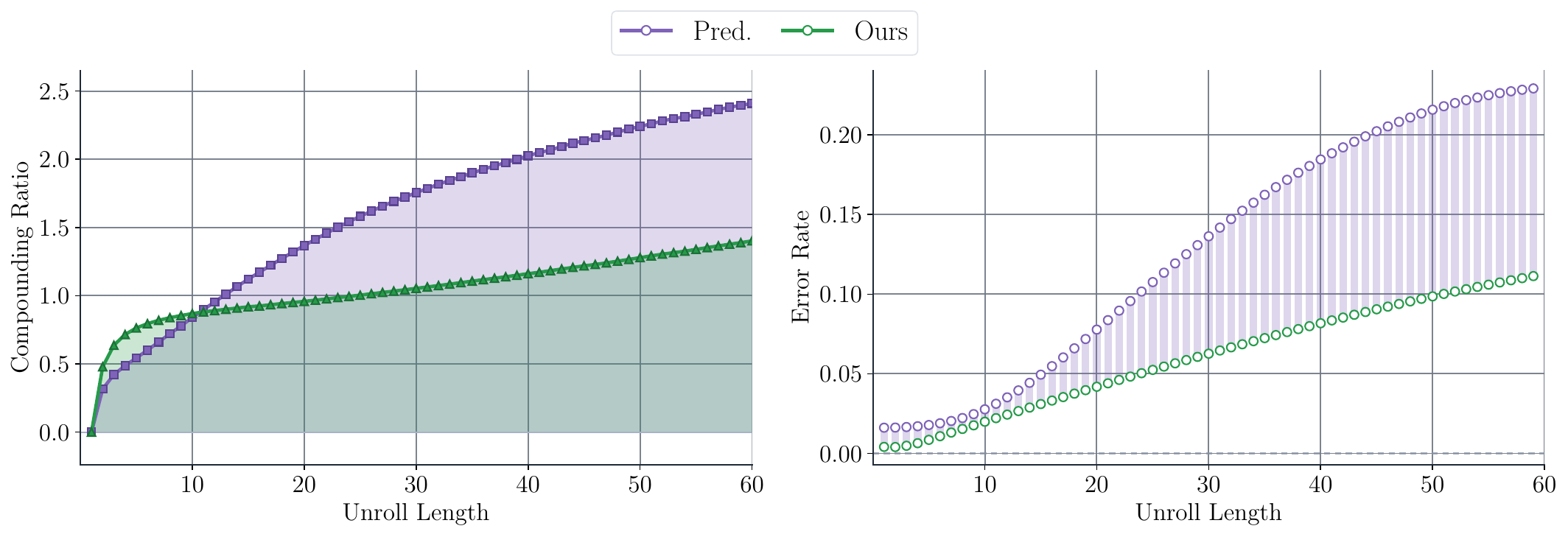}
    \caption{Recursive rollout error analysis. The compounding ratio (left) compares open-loop prediction to teacher-forced prediction; values above $1$ indicate error accumulation caused by recursion. The error growth rate (right) measures the additional pose error introduced at each rollout step. Our method stays closer to teacher-forced behavior and has lower error growth than the predictive baseline, showing that latent-space dynamics modeling mitigates long-horizon compounding error.}
    \label{fig:error_analysis}
    \vspace{-10pt}
\end{figure*}

\begin{figure*}[t]
    \vspace*{4pt} 
    \centering
    \includegraphics[width=\textwidth, trim=0 0 0 0, clip]{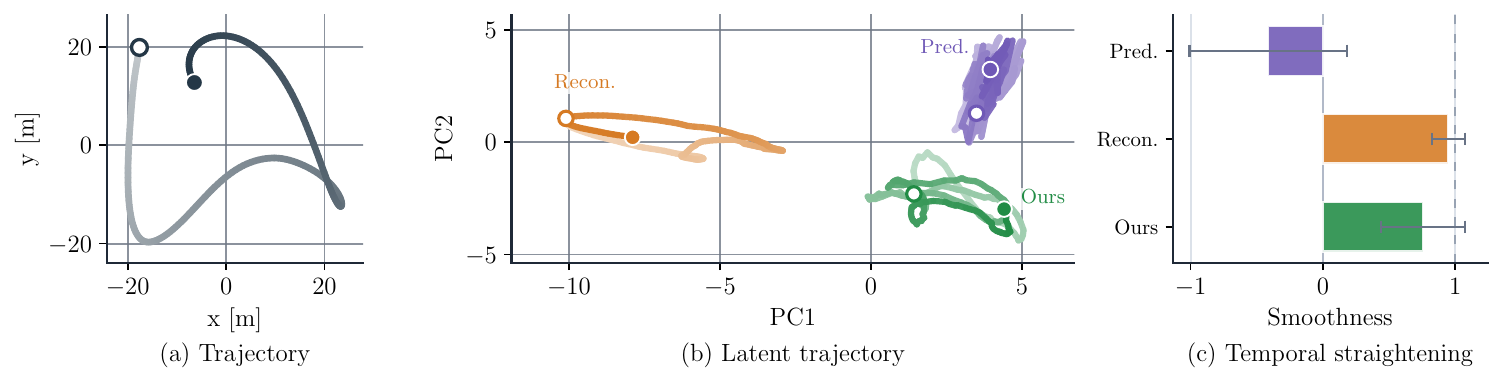}
    \caption{Temporal straightening analysis of latent rollouts. (a) Example Cartesian trajectory, with shading indicating progression from earlier to later states. (b) PCA projection of the corresponding latent trajectories, showing how each model evolves in representation space over time. (c) Temporal straightening score; higher values indicate smoother and more directionally consistent latent evolution. Temporally straighter latent trajectories suggest that complex system dynamics are represented in a simpler geometry, making recursive prediction easier and reducing directional drift during rollout.}
    \label{fig:temporal_straighting}
    \vspace{-10pt}
\end{figure*}

\noindent This metric measures the amount of new error introduced at each rollout step. Lower values indicate slower error accumulation and better long-horizon stability. As shown in Fig.~\ref{fig:error_analysis} on the left, our method has a higher CR than the predictive baseline for the first few rollout steps, particularly for $k<10$. This does not imply that the predictive baseline is better in this regime. At short horizons, both teacher-forced and recursive errors are small, so the ratio can be sensitive to small differences in the denominator of eq.~(\ref{eq::compounding_ratio}). Moreover, values below $1$ indicate that recursive rollout error is still lower than teacher-forced error at that horizon. Therefore, the early-horizon behavior mainly reflects local differences between teacher-forced and rollout trajectories, rather than meaningful long-horizon compounding. The long-horizon trend is more informative. The predictive baseline crosses $\mathrm{CR}_k=1$ around $k\approx 12$ and then grows rapidly, reaching approximately $2.4$ by $k=60$. This indicates that its recursive rollout error becomes more than twice as large as its teacher-forced error. In contrast, our method remains close to $1$ for a much longer portion of the horizon and increases more gradually, reaching only about $1.4$ at $k=60$. This shows that the latent-space model remains closer to its teacher-forced behavior under open-loop rollout. The error growth plot in Fig.~\ref{fig:error_analysis} on the right confirms this interpretation. The predictive baseline introduces larger new error at nearly every rollout step. Around $k\approx 30$, its error growth is approximately $0.14$, while ours is around $0.06$. By the end of the horizon, the predictive baseline reaches about $0.23$, whereas our method remains near $0.11$. Therefore, our method not only has a lower long-horizon compounding ratio, but also injects less new error at each recursive step. These results show that latent-space dynamics modeling substantially mitigates recursive error accumulation compared to direct autoregressive prediction in state space.


\begin{takeawaybox}
\small\textbf{Takeaway:} Latent-space dynamics modeling reduces compounding error and has a much smaller gap between teacher-forced and recursive rollout predictions.
\end{takeawaybox}

\textbf{Temporal Straightening}. We further analyze the geometry of the learned rollouts through the lens of \emph{temporal straightening}, motivated by the hypothesis in~\cite{henaff2019perceptual, wang2026temporal} that complex temporal dynamics can be represented as smoother, straighter trajectories in representation space. The intuition is that a NN model becomes easier to roll out when its internal representation space evolves in a consistent temporal direction: if consecutive latent displacements are well aligned, the trajectory unfolds smoothly over time and future states are easier to extrapolate recursively. In contrast, if the representation repeatedly bends, oscillates, or backtracks, then small local prediction errors can more easily deflect the rollout away from the true temporal progression. Following this idea, we evaluate temporal straightening over \(N\) trajectories. For the \(i\)-th trajectory, let the sequence of predicted latent representations be \(\tilde{s}^{(i)}_{1:T} \in \mathbb{R}^{T \times \bar{D}}\), and define the temporal velocity vectors as \(\dot{\tilde{s}}^{(i)}_t = \tilde{s}^{(i)}_{t+1} - \tilde{s}^{(i)}_t\). The \textit{smoothness score} for a single trajectory is computed as the mean pairwise cosine similarity between consecutive latent velocities
\[
S_{\text{straight}}^{(i)}
=
\frac{1}{T-2}
\sum_{t=1}^{T-2}
\frac{\left\langle \dot{\tilde{s}}^{(i)}_t,\, \dot{\tilde{s}}^{(i)}_{t+1} \right\rangle}
{\|\dot{\tilde{s}}^{(i)}_t\|\,\|\dot{\tilde{s}}^{(i)}_{t+1}\|}.
\]
 We then report the mean and variance across the \(N\) evaluated trajectories. A value close to \(1\) is ideal, since it indicates that successive latent displacements are highly aligned and the rollout follows a smooth, nearly straight path. Values near \(0\) indicate weak alignment and a more curved or wandering trajectory, while negative values indicate strong directional inconsistency, such as oscillation or backtracking. Figure~\ref{fig:temporal_straighting} shows a clear separation between direct predictive modeling and the two latent-space approaches. Averaged over the \(N\) evaluated trajectories, the predictive baseline attains a negative mean straightening score of roughly \(-0.4\), with a large spread extending from about \(-1.0\) to \(0.2\), indicating that its rollouts often change direction and evolve along a temporally inconsistent path. In contrast, both latent models remain strongly in the positive regime: the reconstruction-based latent model achieves the highest mean score, around \(0.95\), while our JEPA-based model attains a mean of about \(0.75\). Thus, despite their different training objectives, both latent models produce substantially smoother and straighter temporal evolution than the predictive baseline. This suggests that an important emergent property of latent dynamics learning is temporal smoothness in representation space, which makes recursive rollouts easier to maintain over long horizons. In our setting, this result helps explain why both latent approaches outperform direct predictive modeling: by organizing temporal evolution into smoother latent trajectories, they reduce directional drift under recursive rollout and are therefore less susceptible to compounding error. While the reconstruction model achieves the highest straightening score, our model remains strongly positive and, when interpreted together with the rollout-error results, indicates that latent-space modeling itself is a key ingredient for learning temporal representations.

\begin{takeawaybox}
\small\textbf{Takeaway:} Both latent models learn temporally smoother latent trajectories than direct predictive modeling, suggesting that temporal smoothness is an important emergent property.
\end{takeawaybox}


\begin{figure*}[t]
    \vspace*{4pt} 
    \centering
    \includegraphics[width=\textwidth, trim=0 0 0 0, clip]{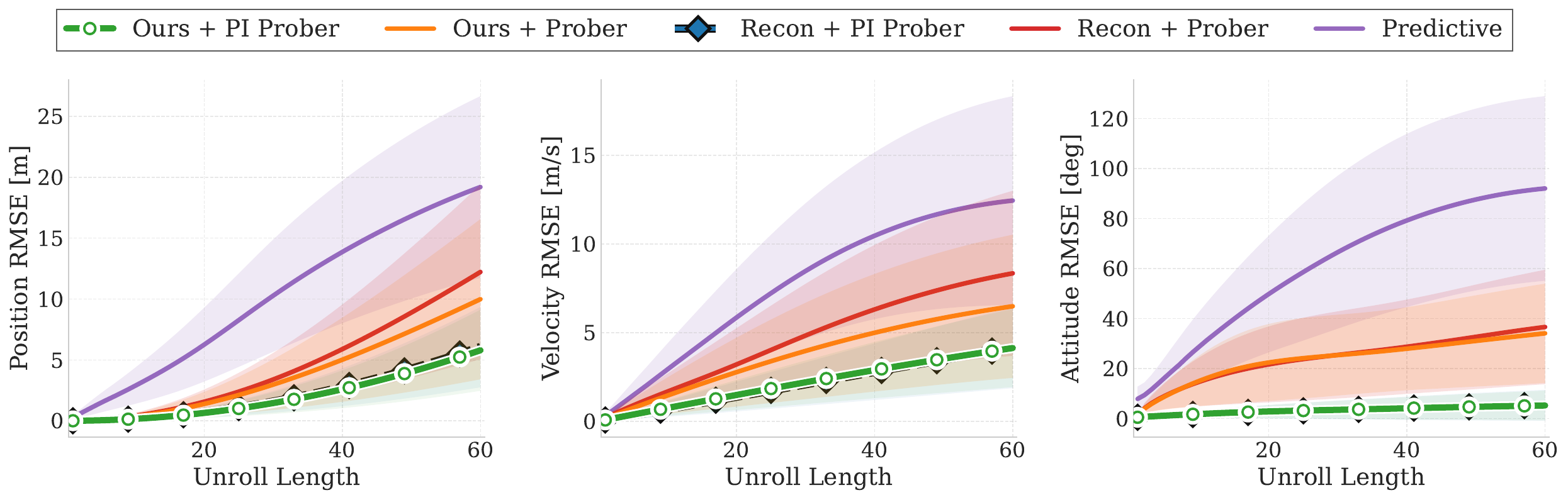}
    \caption{Open-loop rollout fidelity across dynamics models. Position, velocity, and attitude RMSE are shown versus unroll length, where lower error indicates more accurate long-horizon prediction. Direct predictive models accumulate error rapidly, especially in attitude, while latent dynamics models grow more slowly. The proposed physics-inspired prober substantially improves metric-state recovery, showing that structured physical decoding is critical for accurate state prediction.}
    \label{fig:open_loop}
    \vspace{-10pt}
\end{figure*}

\textbf{Robustness to Noise}.
We next evaluate whether the learned dynamics remain reliable under corrupted sensing. To simulate observation noise, we perturb the input observation history with i.i.d. Gaussian noise before passing it to the proposed framework. The corruption is applied only to the model input. The reference ground-truth trajectory remains unchanged for evaluation. We then recursively unroll each model over the prediction horizon and compute the pose RMSE between the predicted trajectory, obtained through the latent dynamics model and physics-inspired prober, and the ground-truth trajectory.

\begin{figure}[!t]
  \centering
  \includegraphics[width=\linewidth]{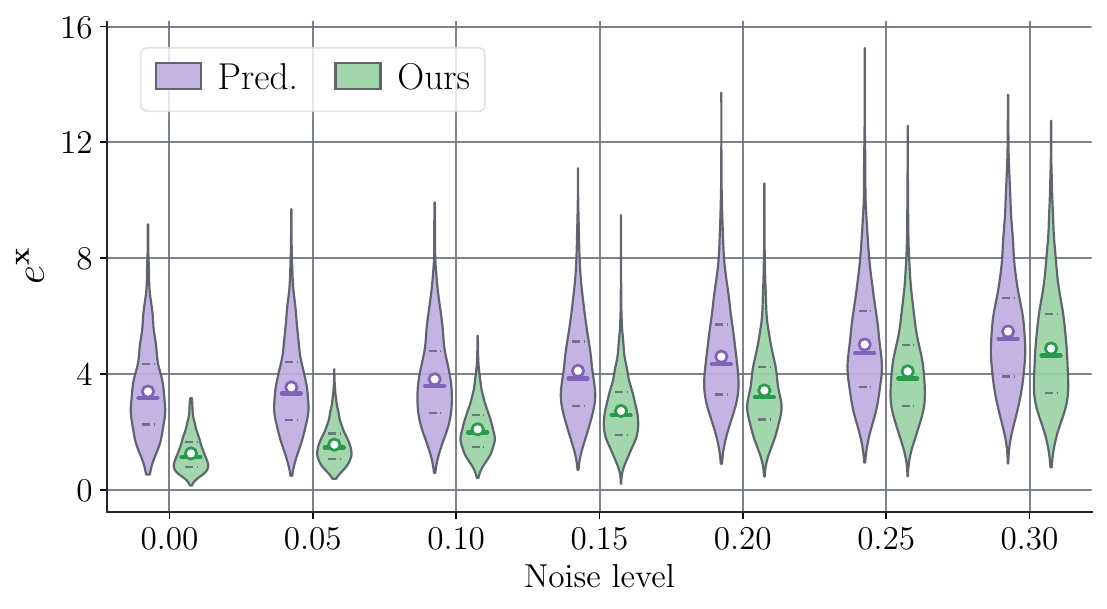}
  \caption{Robustness to observation corruption. Pose RMSE under increasing i.i.d. Gaussian observation noise, showing that the JEPA model consistently outperforms the predictive baseline and remains more robust as corruption increases.}
  \label{fig:noise_ablation}
\end{figure}

Figure~\ref{fig:noise_ablation} shows that our method remains consistently more robust than the predictive baseline as the observation noise increases. 
At zero noise, our method reduces the median state RMSE by approximately $55\%$ compared to the predictive baseline. 
As the noise level increases, this advantage remains clear. 
At moderate noise levels, our method achieves roughly a $25$--$30\%$ reduction in median state RMSE. 
Even at the highest noise level, where both methods degrade, our method still maintains a lower median error, with an improvement of about $10\%$ over the predictive baseline. The distributional trend is also important. The predictive baseline exhibits wider error distributions and heavier upper tails as noise increases, indicating more frequent high-error rollouts. In contrast, our method produces more concentrated distributions across noise levels. This suggests that the latent dynamics model, together with the physics-inspired prober, is less sensitive to input-level perturbations and yields more consistent long-horizon predictions under sensing corruption.

\begin{takeawaybox}
\small\textbf{Takeaway:} Our JEPA-style approach remains more accurate under input corruption, indicating stronger robustness to corrupted inputs.
\end{takeawaybox}

\subsection{Physics Interpretability}

\textbf{Baselines}. To evaluate whether injecting physical structure improves the open-loop fidelity of mapping latent rollouts to metric state space, we compare six baselines. \emph{Predictive} is a direct autoregressive state-space predictor. \emph{Predictive + Physics Reg.} augments this baseline with kinematic physics regularization loss during training. \emph{Recon + Prober} is a latent dynamics model in which the embeddings are trained through a reconstruction: collapse is avoided by forcing the predicted embeddings to decode back the full recorded metric state and the prober is a 3-layed MLP. \emph{Recon + PI Prober} replaces the unconstrained prober in this model with the proposed physics-inspired probing mechanism. \emph{Ours + Prober} is our JEPA-based latent dynamics model with a 3-layered MLP prober. Finally, \emph{Ours + PI Prober} is the full method, combining JEPA latent dynamics with the proposed physics-inspired prober. Together, these baselines isolate three design choices: direct predictive versus latent-space dynamics modeling, reconstruction-based versus JEPA-based embedding learning, and unconstrained versus physics-grounded state.

\begin{table}[t]
\centering
\caption{Open loop prediction analysis of different neural dynamics frameworks.}
\label{tab:open_loop_baselines}
\vspace{1mm}
\renewcommand{\arraystretch}{1.20}
\setlength{\tabcolsep}{6pt}
\begin{tabular}{lcccc}
\toprule\toprule
\multirow{2}{*}{\textbf{Approach}} 
& \multicolumn{2}{c}{\textbf{Pos. RMSE [m]} $\downarrow$}
& \multicolumn{2}{c}{\textbf{Att. Err. [$^\circ$]} $\downarrow$} \\
\cmidrule(lr){2-3} \cmidrule(lr){4-5}
& \textbf{Mean} & \textbf{Var.} & \textbf{Mean} & \textbf{Var.} \\
\midrule
Predictive                                      & $8.80$ & $2.3$ & $53.4$ & $14.9$ \\
Predictive + Physics Reg.~\cite{saviolo2022physics} & $7.12$ & $2.1$ & $49.1$ & $13.2$ \\
Reconstruction + Prober                         & $6.82$ & $1.6$ & $45.2$ & $9.7$ \\
Reconstruction + PI Prober                      & $1.53$ & $0.13$ & $5.28$ & $0.70$ \\
Ours + Prober                                   & $5.56$ & $1.31$ & $40.20$ & $9.30$ \\
\textbf{Ours + PI Prober}                       & $\mathbf{1.43}$ & $\mathbf{0.10}$ & $\mathbf{4.71}$ & $\mathbf{0.50}$ \\
\bottomrule\bottomrule
\end{tabular}
\end{table}

To evaluate the prediction accuracy of each baseline, we decompose the state error into position and attitude components. 
Let $\tilde{\mathbf{p}}_k, \mathbf{p}_k \in \mathbb{R}^3$ denote the predicted and ground-truth positions at time step $k$, respectively, and let 
$\tilde{\mathbf{R}}_k, \mathbf{R}_k \in SO(3)$ denote the corresponding predicted and ground-truth rotation matrices. The position error is defined as
\begin{equation}
e_k^{\mathbf{p}}
=
\left\|
\tilde{\mathbf{p}}_k - \mathbf{p}_k
\right\|_2 .
\label{eq:position_error}
\end{equation}

\noindent The attitude error is computed using the relative rotation between the ground-truth and predicted orientations:
\begin{equation}
\mathbf{R}_{\mathrm{err},k}
=
\mathbf{R}_k^\top \tilde{\mathbf{R}}_k .
\label{eq:relative_rotation_error}
\end{equation}

\noindent We map this relative rotation to the Lie algebra $\mathfrak{so}(3)$ using the logarithmic map:
\begin{equation}
\boldsymbol{\phi}_k
=
\operatorname{Log}_{\mathcal{SO}(3)}
\left(
\mathbf{R}_k^\top \tilde{\mathbf{R}}_k
\right)^\vee
\in \mathbb{R}^3,
\label{eq:attitude_error_vector}
\end{equation}
where $(\cdot)^\vee$ converts a skew-symmetric matrix in $\mathfrak{so}(3)$ to its corresponding vector representation. The scalar attitude error is then given in degrees as
\begin{equation}
e_k^{\mathbf{R}}
=
\frac{180}{\pi}
\left\|
\boldsymbol{\phi}_k
\right\|_2 .
\label{eq:attitude_error}
\end{equation}

\begin{table*}[t]
\centering
\caption{Quantitative results of real-world trajectory tracking across reference trajectories. Each trajectory is executed 5 times. The mean and variance are reported.}
\label{tab:realworld_tracking}
\vspace{1mm}
\renewcommand{\arraystretch}{1.10}
\setlength{\tabcolsep}{3.8pt}
\footnotesize
\resizebox{\textwidth}{!}{%
\begin{tabular}{lcc|cc|cc|cc|cc|cc|cc}
\toprule\toprule
\textbf{Trajectory} 
& $\|\mathbf{v}\|_{\max}$ 
& $\|\dot{\mathbf{v}}\|_{\max}$
& \multicolumn{6}{c|}{\textbf{Position RMSE [m]}}
& \multicolumn{6}{c}{\textbf{Attitude Error [$^\circ$]}} \\
& {[m s$^{-1}$]} & {[m s$^{-2}$]}
& \multicolumn{2}{c|}{\textbf{Ours}}
& \multicolumn{2}{c|}{\textbf{MPPI (Pred.+Phy.)}}
& \multicolumn{2}{c|}{\textbf{MPPI (Pred.)}}
& \multicolumn{2}{c|}{\textbf{Ours}}
& \multicolumn{2}{c|}{\textbf{MPPI (Pred.+Phy.)}}
& \multicolumn{2}{c}{\textbf{MPPI (Pred.)}} \\
\cmidrule(lr){4-9} \cmidrule(lr){10-15}
& &
& \textbf{Mean} & \textbf{Var.}
& \textbf{Mean} & \textbf{Var.}
& \textbf{Mean} & \textbf{Var.}
& \textbf{Mean} & \textbf{Var.}
& \textbf{Mean} & \textbf{Var.}
& \textbf{Mean} & \textbf{Var.} \\
\midrule
Circle     
& $2.45$ & $1.40$
& $\mathbf{0.24}$ & $\mathbf{0.010}$
& $0.36$ & $0.022$
& $0.39$ & $0.024$
& $\mathbf{7.87}$ & $\mathbf{0.82}$
& $10.99$ & $3.66$
& $11.95$ & $3.81$ \\

Oval       
& $4.50$ & $3.78$
& $\mathbf{0.33}$ & $\mathbf{0.018}$
& $0.44$ & $0.083$
& $0.48$ & $0.096$
& $\mathbf{9.11}$ & $\mathbf{1.06}$
& $15.20$ & $5.46$
& $16.53$ & $5.68$ \\

Figure 8   
& $5.20$ & $5.44$
& $\mathbf{0.35}$ & $\mathbf{0.022}$
& $0.47$ & $0.041$
& $0.51$ & $0.044$
& $\mathbf{9.25}$ & $\mathbf{1.18}$
& $17.75$ & $5.88$
& $20.20$ & $6.10$ \\

Fish       
& $5.70$ & $7.68$
& $\mathbf{0.40}$ & $\mathbf{0.031}$
& $0.54$ & $0.056$
& $0.59$ & $0.061$
& $\mathbf{10.78}$ & $\mathbf{1.42}$
& $20.95$ & $6.85$
& $22.78$ & $7.19$ \\

Lemniscate 
& $7.20$ & $12.5$
& $\mathbf{0.45}$ & $\mathbf{0.047}$
& $0.56$ & $0.076$
& $0.61$ & $0.083$
& $\mathbf{19.43}$ & $\mathbf{2.65}$
& $26.83$ & $7.51$
& $29.16$ & $7.90$ \\
\bottomrule\bottomrule
\end{tabular}%
}
\end{table*}

\begin{figure*}[t]
    \vspace*{4pt} 
    \centering
    \includegraphics[width=\textwidth, trim=0 0 0 0, clip]{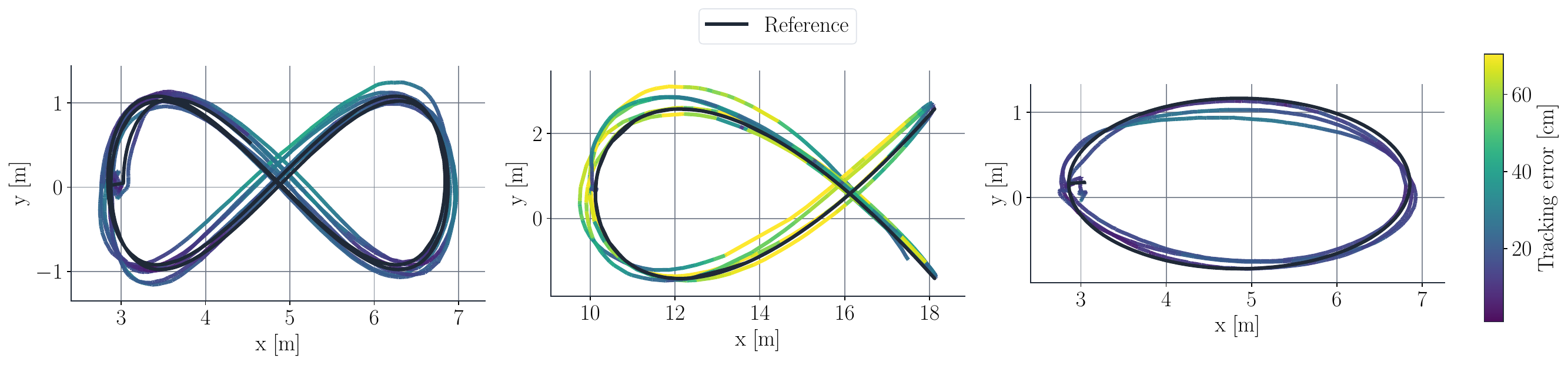}
    \caption{Real-world zero-shot trajectory tracking using the proposed controller. The plots show executed flight trajectories overlaid on the desired references. Color indicates tracking error in centimeters, with darker regions corresponding to lower error. The controller closely follows diverse real-world trajectories using a dynamics model trained only in simulation.}
    \label{fig:traj_tracking}
    \vspace{-10pt}
\end{figure*}

\textbf{Physics Structure in NN Design}. Figure~\ref{fig:open_loop} and Table~\ref{tab:open_loop_baselines} first reveal a clear advantage of latent dynamics models over direct autoregressive prediction. The weakest performance is obtained by the predictive baseline, which reaches \(8.80\) m mean position RMSE and \(53.4^\circ\) mean attitude error, with large rollout variance (\(2.3\) m and \(14.9^\circ\) respectively). Adding kinematic physics regularization improves this baseline only modestly, reducing the mean position RMSE to \(7.12\) m and the mean attitude error to \(49.1^\circ\), corresponding to relative improvements of only \(19\%\) and \(8\%\). This indicates that physics-inspired penalties help, but are insufficient to overcome the underlying instability of autoregressive prediction. In contrast, the latent baselines without PI probing already perform better: \emph{Recon + Prober} achieves \(6.82\) m position RMSE and \(45.2^\circ\) attitude error, while \emph{Ours + Prober} improves further to \(5.56\) m and \(40.2^\circ\). Relative to the predictive baseline, this corresponds to reductions of \(22.5\%\) and \(33.8\%\) in position error, and \(15.4\%\) and \(24.7\%\) in attitude error, for \emph{Recon + Prober} and \emph{Ours + Prober}, respectively. Qualitatively, the rollout curves in Fig.~\ref{fig:open_loop} show the same trend: the predictive variants exhibit rapidly growing position, velocity, and especially attitude error, whereas the latent methods remain consistently lower across the horizon. These results suggest that latent dynamics modeling is already a better design choice than direct autoregressive prediction, and that simply adding physics regularization to a predictive model is not enough.


\textbf{Training Objective}. We next compare the two latent learning paradigms and the effect of injecting kinematic structure into state recovery. Without PI probing, \emph{Ours + Prober} outperforms \emph{Recon + Prober}, reducing mean position RMSE from \(6.82\) to \(5.56\) m and mean attitude error from \(45.2^\circ\) to \(40.2^\circ\). This shows that JEPA-style latent training produces a stronger predictive representation than reconstruction-based training, where embeddings are optimized to reconstruct the recorded state and avoid collapse through decoder supervision. However, the largest gains appear only after introducing the PI Prober. For the reconstruction model, replacing the generic prober with the PI Prober reduces position RMSE from \(6.82\) to \(1.53\) m and attitude error from \(45.2^\circ\) to \(5.28^\circ\), corresponding to roughly \(4.5\times\) and \(8.6\times\) improvements. For our JEPA model, the same replacement reduces position RMSE from \(5.56\) to \(1.43\) m and attitude error from \(40.2^\circ\) to \(4.71^\circ\), i.e., about \(3.9\times\) and \(8.5\times\) improvements. Variance is reduced just as sharply, dropping from \(1.31\) to \(0.10\) in position and from \(9.30^\circ\) to \(0.50^\circ\) in attitude for the JEPA model. The qualitative rollout curves (see Figure~\ref{fig:open_loop}) reinforce the same trend. This suggests that the proposed kinematic structure makes the state-recovery problem substantially easier: instead of learning metric dynamics entirely from scratch through a flexible decoder, the model only needs to predict structured corrections on top of a parameter-free prior. The PI Prober is not learning the full dynamics end-to-end. It rather learns the missing dynamics needed to complement the imposed kinematic structure and recover the full system evolution. Importantly, in our setup this structure can be imposed without requiring nominal parameter estimation. Overall, these results show that JEPA provides the stronger dynamics model, while the PI Prober is the key mechanism that converts embeddings into accurate and stable metric-state trajectories for control.

\begin{takeawaybox}
\small\textbf{Takeaway:} Embedding kinematic structure enables high-fidelity metric-state recovery, showing that the latent rollouts retain sufficient information to reconstruct the full dynamics.
\end{takeawaybox}

\subsection{Zero-Shot Sim-to-Real Transfer}

We now evaluate how far the proposed framework performs when trained entirely on domain-randomized simulation data, with no task-specific fine-tuning on real-world trajectories. We compare three controllers in closed loop: \emph{Ours}, which uses the proposed framework; \emph{MPPI (Predictive + Physics Regularization)}, which uses a direct predictive state-space model trained with physics regularization; and \emph{MPPI (Predictive)}, which uses a standard predictive state-space model without additional physical grounding. These experiments therefore test whether the benefits observed in open-loop prediction translate into real control performance under sim-to-real transfer. Each trajectory is executed 5 times and we report the mean and varience of the tracking performance. Table~\ref{tab:realworld_tracking} shows a consistent advantage of the proposed method across all reference trajectories. In terms of position tracking, \emph{Ours} achieves the lowest mean RMSE on every trajectory, reducing the error from \(0.39\) to \(0.24\) m on the circle (\(\approx 38\%\) improvement over \emph{MPPI (Predictive)}), from \(0.48\) to \(0.33\) m on the oval (\(\approx 31\%\)), from \(0.51\) to \(0.35\) m on the figure-8 (\(\approx 31\%\)), from \(0.59\) to \(0.40\) m on the fish trajectory (\(\approx 32\%\)), and from \(0.61\) to \(0.45\) m on the lemniscate (\(\approx 26\%\)). Similar gains are obtained relative to \emph{MPPI (Predictive + Physics Regularization)}, with position-error reductions ranging from about \(20\%\) to \(33\%\). The same trend appears in attitude tracking. Relative to \emph{MPPI (Predictive)}, our method reduces mean attitude error from \(11.95^\circ\) to \(7.87^\circ\) on the circle (\(\approx 34\%\)), from \(16.53^\circ\) to \(9.11^\circ\) on the oval (\(\approx 45\%\)), from \(20.20^\circ\) to \(9.25^\circ\) on the figure-8 (\(\approx 54\%\)), from \(22.78^\circ\) to \(10.78^\circ\) on the fish trajectory (\(\approx 53\%\)), and from \(29.16^\circ\) to \(19.43^\circ\) on the lemniscate (\(\approx 33\%\)). Variance is also consistently lower for our method in both position and attitude, indicating not only better average tracking but also more repeatable closed-loop behavior across trials.

The qualitative tracking examples in Fig.~\ref{fig:traj_tracking} illustrate representative real-world tracking of the proposed method across multiple reference trajectories. These examples show that the controller is able to track diverse closed-loop paths in the real world using a dynamics model trained entirely in simulation, with the color-coded error remaining generally concentrated along the reference trajectory. The quantitative results in Table~\ref{tab:realworld_tracking} provide the stronger evidence: across all trajectories, our method consistently achieves lower position and attitude errors than both predictive baselines, and this advantage persists as the trajectories become faster and more aggressive. Overall, these results suggest that simulation alone can be sufficient for real-world control when the learned dynamics model is structured appropriately. A likely reason is that the proposed framework combines three complementary ingredients: latent dynamics modeling reduces the compounding effects of direct autoregressive state prediction, domain randomization broadens the training distribution to better capture real-world variability, and the physics-inspired prober grounds latent rollouts in metric state space so that the model remains useful for control. Together, these components enable more reliable zero-shot transfer from simulation to the real platform than purely predictive alternatives.

\begin{table*}[t]
\centering
\caption{Quantitative results of closed-loop tracking under propeller switching and payload scenarios. All trajectories were executed at an average velocity of 2~ms$^{-1}$ and executed 5 times. The table reports the mean and variance.}
\label{tab:robust_tracking}
\vspace{1mm}
\renewcommand{\arraystretch}{1.12}
\setlength{\tabcolsep}{4.2pt}
\footnotesize
\resizebox{\textwidth}{!}{%
\begin{tabular}{ll|cc|cc|cc|cc|cc|cc}
\toprule\toprule
\textbf{Scenario} & \textbf{Trajectory}
& \multicolumn{6}{c|}{\textbf{Position RMSE [m]}}
& \multicolumn{6}{c}{\textbf{Attitude Error [$^\circ$]}} \\
& 
& \multicolumn{2}{c|}{\textbf{Ours}}
& \multicolumn{2}{c|}{\textbf{MPPI (Pred. + Phy.)}}
& \multicolumn{2}{c|}{\textbf{MPPI (Pred.)}}
& \multicolumn{2}{c|}{\textbf{Ours}}
& \multicolumn{2}{c|}{\textbf{MPPI (Pred. + Phy.)}}
& \multicolumn{2}{c}{\textbf{MPPI (Pred.)}} \\
\cmidrule(lr){3-8} \cmidrule(lr){9-14}
& 
& \textbf{Mean} & \textbf{Var.}
& \textbf{Mean} & \textbf{Var.}
& \textbf{Mean} & \textbf{Var.}
& \textbf{Mean} & \textbf{Var.}
& \textbf{Mean} & \textbf{Var.}
& \textbf{Mean} & \textbf{Var.} \\
\midrule

\multirow{3}{*}{\textbf{Propeller Switching}}
& Circle
& $\mathbf{0.33}$ & $\mathbf{0.02}$
& $0.43$ & $0.03$
& $0.45$ & $0.03$
& $\mathbf{9.54}$ & $\mathbf{1.04}$
& $12.40$ & $1.35$
& $12.88$ & $1.40$ \\

& Figure 8
& $\mathbf{0.35}$ & $\mathbf{0.01}$
& $0.46$ & $0.01$
& $0.47$ & $0.01$
& $\mathbf{10.21}$ & $\mathbf{0.98}$
& $13.27$ & $1.27$
& $13.78$ & $1.32$ \\

& Fish
& $\mathbf{0.39}$ & $\mathbf{0.02}$
& $0.51$ & $0.03$
& $0.53$ & $0.03$
& $\mathbf{10.89}$ & $\mathbf{0.88}$
& $14.16$ & $1.14$
& $14.70$ & $1.19$ \\

\midrule

\multirow{3}{*}{\textbf{Payload Transportation}}
& Circle
& $\mathbf{0.46}$ & $\mathbf{0.08}$
& $0.60$ & $0.10$
& $0.62$ & $0.11$
& $\mathbf{10.11}$ & $\mathbf{1.43}$
& $13.14$ & $1.86$
& $13.65$ & $1.93$ \\

& Figure 8
& $\mathbf{0.49}$ & $\mathbf{0.07}$
& $0.64$ & $0.09$
& $0.66$ & $0.09$
& $\mathbf{9.44}$ & $\mathbf{1.37}$
& $12.27$ & $1.78$
& $12.74$ & $1.85$ \\

& Fish
& $\mathbf{0.53}$ & $\mathbf{0.08}$
& $0.69$ & $0.10$
& $0.72$ & $0.11$
& $\mathbf{11.87}$ & $\mathbf{1.92}$
& $15.43$ & $2.50$
& $16.02$ & $2.59$ \\

\bottomrule\bottomrule
\end{tabular}%
}
\end{table*}

\begin{figure*}[t]
    \vspace*{4pt} 
    \centering
    \includegraphics[width=\textwidth, trim=0 0 0 0, clip]{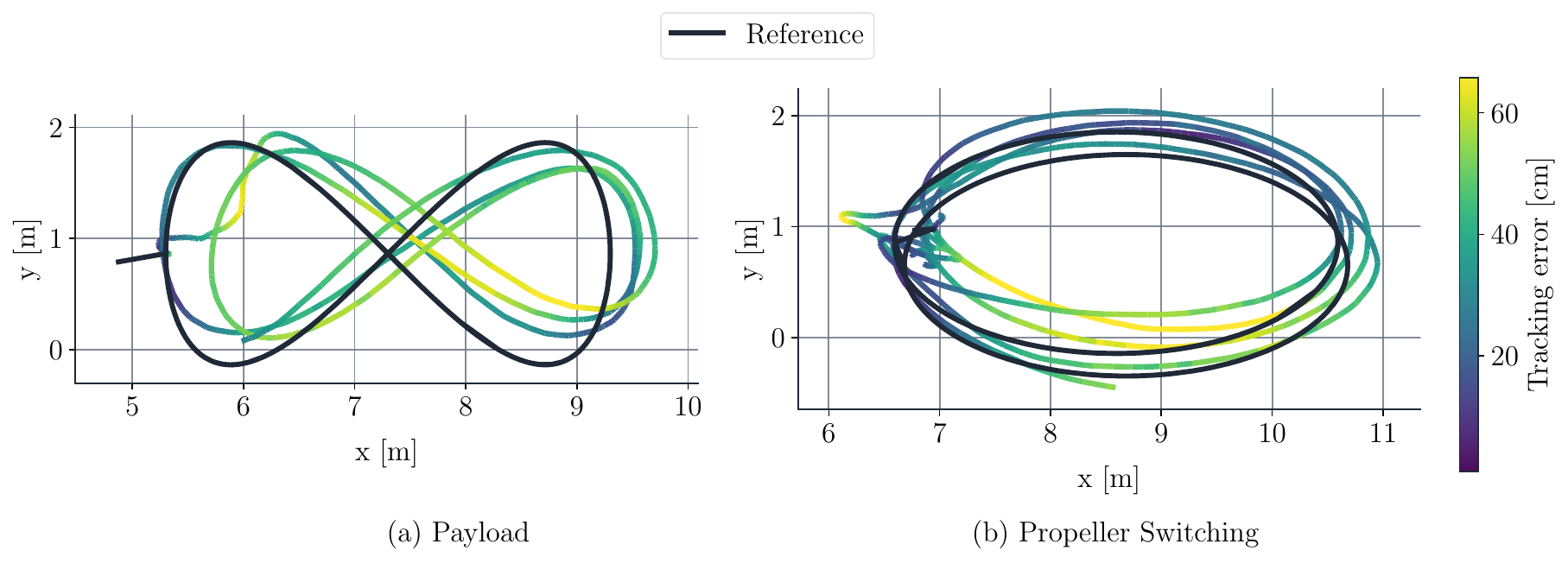}
    \caption{Real-world tracking under platform changes without retraining. (a) Payload transportation modifies the mass and inertial properties of the quadrotor. (b) Propeller switching changes the actuation characteristics. The executed trajectories remain close to the reference paths, with color indicating tracking error in centimeters. These rollouts show that the proposed framework remains effective under non-nominal deployment conditions.}
    \label{fig:geometric_changes}
    \vspace{-10pt}
\end{figure*}

\subsection{Robustness}
We finally evaluate whether the proposed framework remains effective when the real platform deviates from its nominal training configuration. In particular, we consider two deployment scenarios that alter the vehicle dynamics without any retraining or task-specific adaptation: \emph{propeller switching}, which changes the actuation characteristics, and \emph{payload transportation}, which modifies the vehicle mass and inertial properties. These scenarios were not covered during data collection. These experiments are significant because such changes are common in real deployments and can induce substantial model mismatch. All trajectories were executed at an average velocity of 2~ms$^{-1}$ and executed 5 times. We compare the same three closed-loop controllers as before: \emph{Ours}, \emph{MPPI (Pred.+Phy.)}, and \emph{MPPI (Pred.)}. The goal is to assess whether latent-space dynamics modeling yields a controller that is more robust to platform variations than direct predictive baselines.

Table~\ref{tab:robust_tracking} shows that our method achieves the best performance across both deployment scenarios. Under \emph{propeller switching}, our method consistently attains about \(1.3\times\) lower position RMSE than \emph{MPPI (Pred.)} and about \(1.2\times\) lower error than \emph{MPPI (Pred.+Phy.)} across the evaluated trajectories. A similar trend appears in attitude tracking, where our method delivers roughly \(1.3\times\) lower error than the predictive baseline and about \(1.25\times\) lower error than the physics-regularized variant. Under \emph{payload transportation}, with a payload of $300g$ the same pattern persists. Our method again achieves approximately \(1.35\times\) lower position RMSE than \emph{MPPI (Pred.)} and about \(1.27\times\) lower error than \emph{MPPI (Pred.+Phy.)}. Attitude error follows a nearly identical trend, with our method consistently providing about \(1.4\times\) lower error than the predictive baseline. Variance is also lower across both position and attitude metrics, indicating not only improved tracking accuracy but also more repeatable behavior under model mismatch.

The qualitative examples in Fig.~\ref{fig:geometric_changes} illustrate representative tracking rollouts of our method under both propeller switching and payload attachment. Even under these non-nominal conditions, the controller remains close to the reference path, with tracking error remaining concentrated along the trajectory rather than diverging in curved segments. When interpreted together with the quantitative results, these experiments suggest that the proposed latent dynamics model is more robust than predictive models because it captures a more compact and transferable representation of system evolution, rather than depending directly on precise next-state prediction under a fixed nominal configuration. As a result, when the platform dynamics shift due to changes in actuation or mass properties, the learned latent dynamics appear to degrade more gracefully than direct predictive state-space models.

\begin{takeawaybox}
\small\textbf{Takeaway:} A well covered diverse domain-randomized simulation data is sufficient for zero-shot sim2real deployment.
\end{takeawaybox}

\subsection{Data Quality}
\textbf{Trajectory Distribution Quality.}
The performance of a learned dynamics model depends strongly on the distribution of transitions used for training (see Section~\ref{sec:data_collection}). We therefore introduce a Trajectory Distribution Quality (TDQ) score to quantify whether $\mathcal{D}$ provides broad, dynamically informative, and robust coverage for learning the transition map. Since the relevant variables are continuous and high-dimensional, measuring coverage of the state--action space is difficult. Therefore, we discretize each normalized feature space using clustering, which provides an empirical partition of the data distribution into local regions. The occupancy of these regions can then be used to compute an entropy-based measure of distributional spread~\cite{shannon1948mathematical,cover1999elements}, while the number of occupied regions measures coverage. This clustering-based discretization is similar in spirit to vector quantization and $K$-means-based density summarization~\cite{lloyd1982least,macqueen1967some}. TDQ measures three properties that directly affect neural dynamics learning: state--action coverage, transition richness, and parameter robustness.

Each trajectory in the dataset is obtained using a simulator parameter vector $\boldsymbol{\eta}$ (see eq.~(\ref{eq:sim_parameter_set})), which specifies the physical parameters of the randomized quadrotor model. For each transition $i$, we define the augmented state--action feature vector
\begin{equation}
    \mathbf{y}_{i}
    =
    \begin{bmatrix}
        \mathbf{v}_{i}^{\top} &
        \mathbf{r}_{x,i}^{\top} &
        \mathbf{r}_{y,i}^{\top} &
        \mathbf{r}_{z,i}^{\top} &
        \boldsymbol{\omega}_{i}^{\top} &
        \mathbf{a}_{i}^{\top}
    \end{bmatrix}^{\top},
\label{eq:tdq_state_action_feature}
\end{equation}
which includes the velocity, attitude, angular velocity, and motor forces. We omit absolute position from $\mathbf{y}_{i}$ because the local quadrotor dynamics are primarily governed by velocity, attitude, angular velocity, and control input. To measure the diversity of local transitions, we define
\begin{equation}
    \mathbf{g}_{i}
    =
    \begin{bmatrix}
        \mathbf{y}_{i}^{\top} &
        \Delta \mathbf{y}_{i}^{\top}
    \end{bmatrix}^{\top},
    \qquad
    \Delta \mathbf{y}_{i}
    =
    \mathbf{y}_{i+1}
    -
    \mathbf{y}_{i}.
\label{eq:tdq_transition_feature}
\end{equation}
The feature $\mathbf{y}_{i}$ measures which regions of the state--action space are visited, while $\mathbf{g}_{i}$ measures how the system locally evolves from those regions. 

\begin{figure}[!t]
  \centering
  \includegraphics[width=\linewidth]{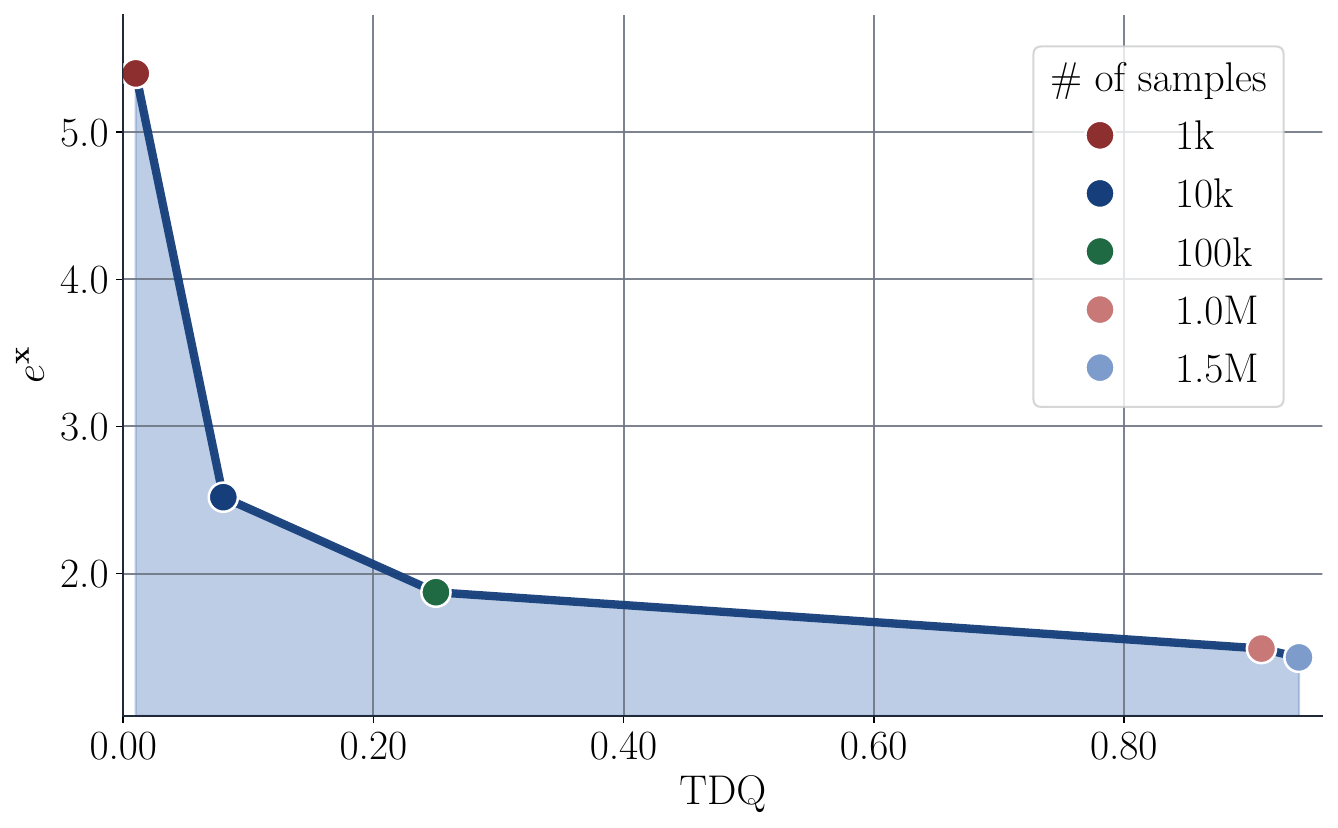}
  \caption{As dataset size increases, TDQ rises from \(0.01\) to \(0.94\), while state RMSE \(e^{\mathbf{x}}\) decreases from \(5.4\) to \(1.4\), showing that higher trajectory distribution quality correlates with better learned dynamics prediction.}
  \label{fig:tdq}
\end{figure}

After normalizing each feature space, we cluster the state--action features $\mathbf{y}_{i}$, transition features $\mathbf{g}_{i}$, and simulator parameter vectors $\boldsymbol{\eta}$ using $K_y$, $K_g$, and $K_\eta$ clusters, respectively. For a generic clustered feature space with $K$ clusters, let $q_k$ denote the empirical occupancy probability of cluster $k$, and let $n_k$ denote the number of samples assigned to that cluster. We define the entropy--coverage score
\begin{equation}
    S
    =
    \left(
    -\frac{1}{\log K}
    \sum_{k=1}^{K}
    q_k \log(q_k)
    \right)
    \left(
    \frac{1}{K}
    \sum_{k=1}^{K}
    \mathbb{I}[n_k \geq n_{\min}]
    \right),
\label{eq:tdq_entropy_coverage_score}
\end{equation}
where the first term is a normalized Shannon entropy that measures how uniformly the data occupies the clustered feature space, and the second term measures the fraction of clusters that are sufficiently represented. The threshold $n_{\min}$ prevents isolated or rarely visited regions from being counted as meaningfully covered. Applying eq.~(\ref{eq:tdq_entropy_coverage_score}) to $\mathbf{y}_{i}$, $\mathbf{g}_{i}$, and $\boldsymbol{\eta}$ gives the state--action coverage score $S_{\mathrm{cov}}$, transition richness score $S_{\mathrm{dyn}}$, and parameter robustness score $S_{\eta}$, respectively. The final TDQ score is defined as the harmonic mean
\begin{equation}
    \mathrm{TDQ}(\mathcal{D})
    =
    \frac{3}
    {
    \frac{1}{S_{\mathrm{cov}}}
    +
    \frac{1}{S_{\mathrm{dyn}}}
    +
    \frac{1}{S_{\eta}}
    }.
\label{eq:tdq_score}
\end{equation}
The harmonic mean penalizes datasets that score poorly in any one component, ensuring that a high TDQ value requires broad state--action coverage, diverse local transition behavior, and sufficient coverage of the randomized simulator parameters. 

To evaluate whether the proposed TDQ score reflects the usefulness of a dataset for dynamics learning, we train the proposed latent dynamics framework using datasets of increasing size and compute the resulting state prediction error \(e^{\mathbf{x}}\). For each dataset size, we compute TDQ from the corresponding training distribution and evaluate the trained model on the same held-out test set. Figure~\ref{fig:tdq} shows a clear inverse relationship between TDQ and predictive error: as TDQ increases, the state RMSE \(e^{\mathbf{x}}\) decreases. With only \(1\mathrm{k}\) samples, the dataset has a very low TDQ of approximately \(0.01\), and the model produces a large state error of about \(5.4\). Increasing the dataset to \(10\mathrm{k}\) samples improves TDQ to roughly \(0.08\), reducing \(e^{\mathbf{x}}\) to about \(2.5\). With \(100\mathrm{k}\) samples, TDQ further increases to approximately \(0.25\), and the error decreases to about \(1.9\), indicating that additional data substantially improves state--action coverage, transition richness, and parameter robustness. Beyond this point, the curve begins to flatten: increasing the dataset size to \(1.0\mathrm{M}\) and \(1.5\mathrm{M}\) raises TDQ to approximately \(0.91\) and \(0.94\), respectively, but only reduces \(e^{\mathbf{x}}\) from about \(1.5\) to \(1.4\). This saturation suggests that the dataset increasingly covers the relevant dynamics manifold, and that further gains are likely limited by model capacity, optimization, or residual dynamics complexity rather than data coverage alone.

\begin{takeawaybox}
\small\textbf{Takeaway:} Higher TDQ leads to lower state prediction error, indicating that broader state--action coverage, richer transitions, and better parameter diversity improve learned dynamics prediction.

\end{takeawaybox}

\section{Discussion}
The experimental results demonstrate that the proposed framework satisfies the four key properties of a useful world model for quadrotor control. First, the model improves \textit{long-horizon prediction} by reducing recursive error accumulation, as shown by the compounding ratio and error growth analysis in Fig.~\ref{fig:error_analysis}. Second, the physics-inspired prober makes the latent rollout \textit{interpretable} by mapping learned representations back to physically meaningful state variables, which is essential for model-based control. This is supported by the open-loop prediction results in Fig.~\ref{fig:open_loop} and Table~\ref{tab:open_loop_baselines}, where the proposed PI Prober substantially improves metric-state recovery. Third, the resulting model is \textit{real-time} for real-time control, since it is directly used inside an MPPI controller onboard an embedded device. Finally, the model can \textit{zero-shot generalize} across tasks: the same learned model is used across multiple trajectories and deployment conditions without task-specific retraining or fine-tuning, as shown in Tables~\ref{tab:realworld_tracking} and~\ref{tab:robust_tracking}.

A central result of this work is that a well-designed simulation data pipeline, combined with an appropriate learning framework, can be sufficient for zero-shot sim-to-real transfer. Rather than relying on accurate system identification or task-specific fine-tuning, our approach systematically collects diverse trajectories and exposes the model to a broad family of plausible quadrotor dynamics through domain randomization, as summarized in Fig.~\ref{fig:tdq}. The TDQ analysis in Fig.~\ref{fig:tdq} further supports this design choice by showing that higher-quality training distributions consistently lead to lower state prediction error. This indicates that broader state--action coverage, richer transition behavior, and greater parameter diversity improve the usefulness of the dataset for learning dynamics. The improvement is most pronounced when moving from small, poorly covered datasets to larger datasets that better span the relevant dynamics distribution; as coverage increases, the gains begin to saturate, suggesting that further improvements are likely limited by model capacity, optimization, or residual dynamics complexity. The real-world tracking results in Table~\ref{tab:realworld_tracking} show that this data-generation strategy is sufficient to outperform predictive baselines across diverse reference trajectories. More importantly, the robustness experiments in Table~\ref{tab:robust_tracking} show that the same model remains effective under deployment changes such as propeller switching and payload transportation. This suggests that the learned representation captures dynamics that are not tied to a single nominal platform. While this work focuses on simulation-only training, the same framework could also be applied directly to real-world data to capture difficult-to-model nonlinear effects, especially at high speeds where aerodynamic and actuator effects become more pronounced.

Another practical advantage of the proposed framework is the simplicity of the SIGReg training objective. Many representation learning methods require multiple carefully tuned loss terms~\cite{bardes2021vicreg,sobal2025learning}, target networks, stop-gradient choices~\cite{grill2020bootstrap}, or reconstruction weights. In contrast, SIGReg introduces a compact anti-collapse mechanism with only one main regularization weight, $\lambda_{\mathrm{sig}}$, to tune. This makes the objective easier to use in practice and more appealing for general-purpose dynamics representation learning. Combined with the physics-inspired prober, this leads to a framework that is both expressive and structured: the latent model learns temporally coherent predictive representations, while the prober converts them into interpretable rollouts suitable for control.

\section{Conclusion and Future Works}
In this work, we introduced the first JEPA-style latent dynamics framework for real-time quadrotor control. The proposed approach learns predictive representations of system evolution without directly reconstructing future states, and uses a physics-inspired prober to map frozen latent rollouts into physically meaningful metric states. This design addresses the four key requirements of a useful world model for aerial control. It improves long-horizon prediction by reducing recursive error accumulation, provides interpretable state-space rollouts for evaluating costs and constraints, remains computationally practical for real-time MPPI control, and is task-agnostic across trajectories and deployment conditions. To reduce reliance on costly and risky real-world data collection, we also developed a domain-randomized simulation pipeline for automated dataset generation. Across open-loop prediction, noise robustness, zero-shot sim-to-real transfer, and platform-variation experiments, the proposed framework consistently outperformed direct predictive baselines and demonstrated robust real-world deployment without task-specific fine-tuning.

Future works will extend this framework from low-dimensional state inputs to high-dimensional observations such as RGB and RGB-D images. This setting is a natural fit for JEPA-style learning, since direct predictive reconstruction becomes increasingly expensive and often forces the model to preserve task-irrelevant visual details. Such an extension would open the door to a broad class of visual navigation tasks, including goal-directed flight from onboard camera observations, obstacle avoidance in cluttered environments, and navigation toward semantic targets. This direction is especially important for quadrotors, where observations compared to other domains like in robot manipulation, are egocentric and tightly coupled to the vehicle's underactuated dynamics. A second direction is to incorporate safety more explicitly into the learned representation and planning objective. We would like to investigate how latent dynamics models can encode safety-relevant structures, such as obstacle proximity, visibility, uncertainty, and recoverability, to support reliable planning in cluttered and uncertain environments.

\bibliographystyle{IEEEtran}
\bibliography{references}

\end{document}